\documentclass{article} 
\usepackage{iclr2024_conference,times}

\usepackage{amsmath,amsfonts,bm}

\def\eqref#1{equation~\ref{#1}}

\def\1{\bm{1}}

\DeclareMathAlphabet{\mathsfit}{\encodingdefault}{\sfdefault}{m}{sl}
\SetMathAlphabet{\mathsfit}{bold}{\encodingdefault}{\sfdefault}{bx}{n}

\usepackage{hyperref}
\usepackage{graphicx}
\usepackage{url}
\usepackage{algorithm}
\usepackage{algorithmic}
\usepackage{newfloat}
\usepackage{listings}
\usepackage{amsmath}
\usepackage{amssymb}
\usepackage{subcaption}
\usepackage{caption}
\usepackage{booktabs}
\usepackage{multirow}
\usepackage{xcolor}
\usepackage{makecell}
\usepackage{xspace}
\usepackage{xcolor,colortbl}
\usepackage{enumitem}

\newcommand{\ie}{\textit{i.e.\xspace}}

\definecolor{Gray}{rgb}{0.9255,0.9255,0.9176}
\definecolor{LightCyan}{rgb}{0.7569,0.7961,0.8431}
\newcommand{\modelname}{\textit{OpenLEAF}\xspace}

\newcommand{\bingname}{BingChat\xspace}

\title{OpenLEAF: Open-Domain Interleaved Image-Text Generation and Evaluation}

\author{Jie An$^1$\thanks{Work done during internship at Microsoft.} , ~~Zhengyuan Yang$^2$, ~~Linjie Li$^2$, ~~Jianfeng Wang$^2$, ~~Kevin Lin$^2$ \\ 
\textbf{Zicheng Liu$^2$, ~~Lijuan Wang$^2$, ~~Jiebo Luo$^1$} \\
    $^1$University of Rochester,\quad
    $^2$Microsoft Azure AI\\%
    \tt\small{\{jan6,jluo\}@cs.rochester.edu},\\ 
    \tt\small\{zhengyang,lindsey.li,jianfw,keli,zliu,lijuanw\}@microsoft.com
}

\iclrfinalcopy 
\begin{document}

\maketitle
\pagestyle{plain}

\begin{abstract}

This work investigates a challenging task named open-domain interleaved image-text generation, which generates interleaved texts and images following an input query. We propose a new interleaved generation framework based on prompting large-language models (LLMs) and pre-trained text-to-image (T2I) models, namely \modelname. In \modelname, the LLM generates textual descriptions, coordinates T2I models, creates visual prompts for generating images, and incorporates global contexts into the T2I models. 
This global context improves the entity and style consistencies of images in the interleaved generation.
For model assessment, we first propose to use large multi-modal models (LMMs) to evaluate the entity and style consistencies of open-domain interleaved image-text sequences.
According to the LMM evaluation on our constructed evaluation set, the proposed interleaved generation framework can generate high-quality image-text content for various domains and applications, such as how-to question answering, storytelling, graphical story rewriting, and webpage/poster generation tasks. Moreover, we validate the effectiveness of the proposed LMM evaluation technique with human assessment. 
We hope our proposed framework, benchmark, and LMM evaluation could help establish the intriguing interleaved image-text generation task.

\end{abstract}
\begin{figure}[t]
    \centering
    \includegraphics[width=.48\textwidth]{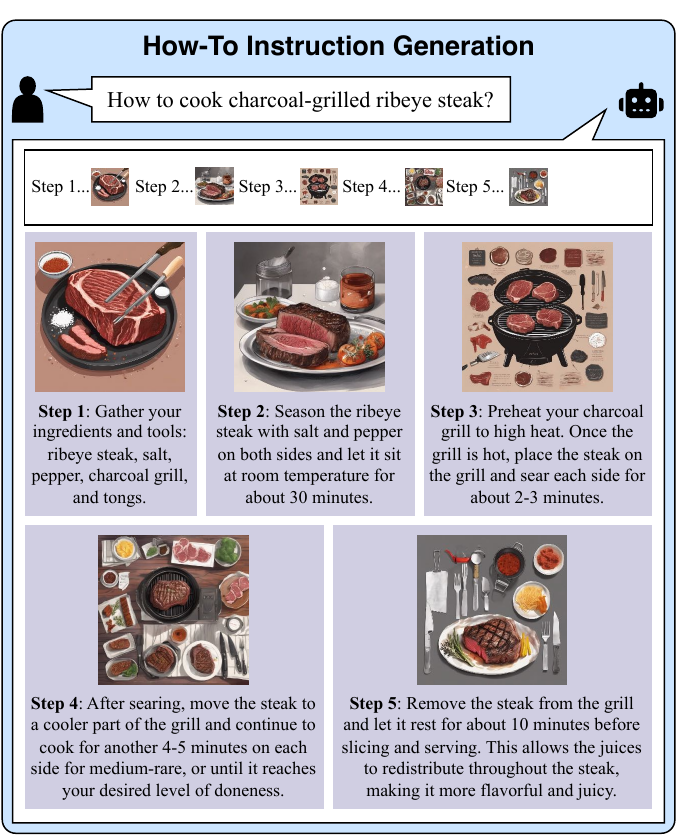}
    \includegraphics[width=.48\textwidth]{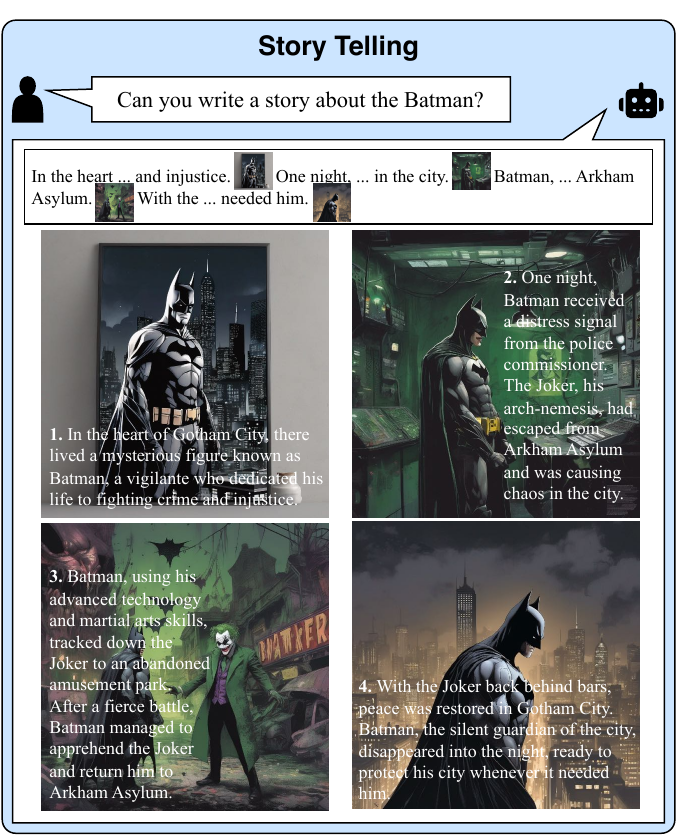}
    \includegraphics[width=.48\textwidth]{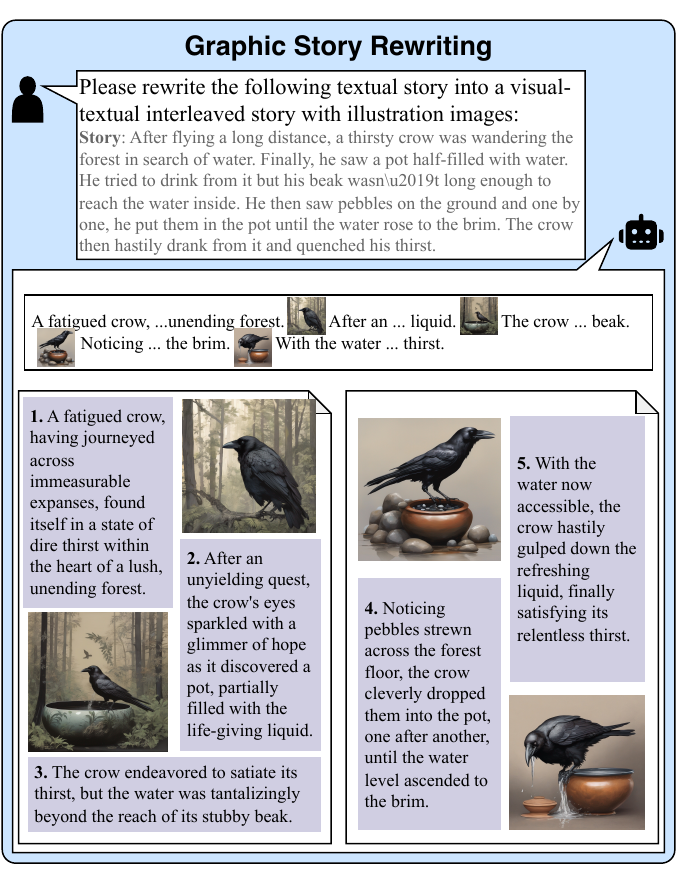}
    \includegraphics[width=.48\textwidth]{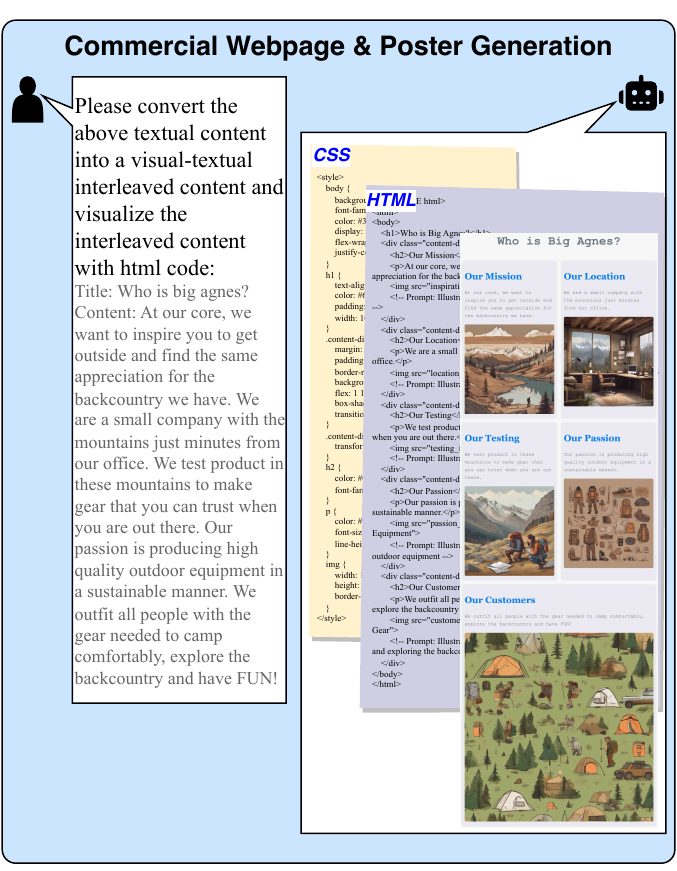}
    \vspace{-2mm}
    \caption{Examples of open-domain interleaved content generation. We show baseline results on producing visual how-to instructions (top-left), generating multi-modal stories (top-right), converting textual stories to multi-modal stories (bottom-left), and generating webpages and posters via HTML and CSS codes (bottom-right).}
    \vspace{-4mm}
    
    \label{fig:intro}
\end{figure}

\section{Introduction}
\label{sec:intro}
This work investigates an intriguing yet challenging task, namely open-domain interleaved image-text generation.
As shown in Fig.~\ref{fig:intro}, given an arbitrary open-domain query as the input, the task aims to generate a sequence of interleaved text descriptions and illustration images to form a coherent content following the input query. 
The ultimate goal of open-domain interleaved generation is to seamlessly generate arbitrary multi-modal contents, thus facilitating a wide range of applications and functionalities, such as generating multi-modal illustrations and web pages, storytelling, chain-of-thought explanations, and so on. 

Early explorations~\citep{li2019storygan,zeng2019pororogan,li2020improved,song2020character,maharana2021improving,maharana2021integrating,maharana2022storydall,szHucs2022modular,pan2022synthesizing,liu2023intelligent} simplify interleaved generation by narrowing down the problem to specific sub-domains, such as the story telling as shown in Fig.~\ref{fig:intro}. 
Despite the promising explorations, these methods can only generate story illustrations in a format in which each image is paired with one single sentence. Therefore, these methods cannot achieve open-domain interleaved generation with arbitrarily interleaved content~\citep{zhu2023multimodal,gadre2023datacomp,laurenccon2023obelisc}, where image and text are organized in diverse interleaved formats for input instructions covering a wide range of open-domain topics. 

For the interleaved content evaluation, prior studies typically train a distinct evaluator for each aspect within a specific domain, such as a story character recognizer~\citep{li2019storygan,maharana2021improving} for character consistency in story generation. 
However, these isolated evaluators cannot scale to open-domain interleaved generation due to the difficulty in generalizing to open-domain generative scenarios.
Consequently, the remaining challenges of the open-domain interleaved generation are the lack of a proper evaluation approach for varying topics and diverse formats, a solid baseline, and a benchmark dataset to compare different methods on.

To address the lack of a baseline model, we propose a new training-free open-domain interleaved image-text generation framework based on GPT-4~\citep{openai2023gpt4} and Stable Diffusion XL (SDXL)~\citep{podell2023sdxl}, named~\modelname. 
Given an arbitrary user query, \modelname first prompts GPT-4 to generate a sequence of textual descriptions, including image placeholders. 
Next, we use GPT-4 to generate T2I prompts that fit well in the interleaved content. 
To improve the entity and style consistency of images within the interleaved content, we add global entity and style contexts to T2I prompts, where the entity context is short appearance descriptions of the common subjects throughout the interleaved content while the style context defines the target image style, both are produced by prompting GPT-4.
Finally, SDXL converts T2I prompts into images, completing the interleaved content by replacing each image tag with the corresponding image.
 
For the interleaved content evaluation, inspired by the power of recent large multi-modal models (LMMs)~\citep{openai2023gpt4,gpt4v,bard,bingchat,yang2023dawn}, we explore prompting LMMs for unified interleaved content evaluation.
Specifically, We use \bingname for the evaluation due to the empirical observation that it performs the best in assessing the quality of an interleaved image-text sequence from different perspectives.
Ideally, an interleaved sequence should contain relevant and informative content, maintain consistency in subject identity and appearance, and feature a coherent image style. 
Therefore, we examine the entity and style consistency of the generated sequence, each broken down into sub-topics for LMM assessment. We then aggregate the generated scores as the final rating.

We collect a benchmark dataset for evaluating open-domain interleaved generation methods, which consists of $30$ input queries, covering a wide range of topics and formats, such as visual instruction generation, story generation and rewriting, webpage, and poster generation. The experiments on this new benchmark based on the LMM-based evaluation approach demonstrate that the proposed baseline method can produce good interleaved results, achieving improvement against a simplified variant. In addition, we design comprehensive human evaluation and LMM analysis on the benchmark to validate the effectiveness of the \bingname evaluation. 

Our contributions are summarized as follows.
\vspace{-3pt}
\begin{itemize}
\setlength\itemsep{-1pt}
    \item We study the open-domain interleaved image-text generation task, which aims to synthesize open-domain interleaved image-text sequences with arbitrary formats.
    \item We first explore LMM's assessment ability to evaluate interleaved image-text sequences. Comprehensive human evaluation and analysis validate the effectiveness of the LMM-based evaluation technique.
    \item We propose \modelname framework as a strong baseline for the interleaved image-text generation task, which improves the semantic consistency and style consistency with the proposed global context. We benchmark \modelname on the developed evaluation set that covers diverse topics and sequence formats.
\end{itemize}
\section{Related Work}
\noindent\textbf{Interleaved Image-Text Generation.}
Starting from the GAN-based methods~\citep{li2019storygan,zeng2019pororogan,li2020improved,song2020character,maharana2021improving,maharana2021integrating,szHucs2022modular}, current methods that address the interleaved image-text generation task include StoryGen~\citep{liu2023intelligent}, AR-LDM~\citep{pan2022synthesizing}, and StoryDALL-E~\citep{maharana2022storydall}, where~\citet{liu2023intelligent} and \citet{pan2022synthesizing} fine-tune latent diffusion models (LDMs)~\citep{rombach2022high} on sequential story-like images, leading to auto-regressive LDMs, while \citet{maharana2022storydall} fine-tunes a text-to-image transformer. However, all of the above methods cannot be applied to the open-domain interleaved generation since the fine-tuning process makes the above model only work well on images similar to the fine-tuning datasets, leading to limited generation domains. On the other hand, how to evaluate open-domain interleaved content remains to be unsettled. 

Recent multi-modal LLMs such as GILL~\citep{koh2023generating}, Emu~\citep{sun2023generative}, and DreamLLM~\citep{dong2023dreamllm} show the decent performance on open-domain image-text generations and perceptions. However, they are not specifically designed for interleaved generation and evaluation, leaving the open-domain interleaved generation task unestablished.

\noindent\textbf{Foundation Models for Open-Domain Evaluation.}
How to evaluate open-domain content has drawn increasing attention. In the natural language process, studies show promises of prompting LLMs such as GPT for open-ended text evalaution~\citep{chiang2023can,liu2023gpteval,fu2023gptscore}.
For evaluating the visual-language content, CLIPscore~\citep{radford2021learning}, VisualGPTScore~\citep{lin2023visualgptscore}, and LLaVA-based scoring methods~\citep{black2023training,liu2023visual} effectively evaluates the open-domain image-text similarity. However, these methods can only work on single image-text pairs while having limited capabilities to evaluate arbitrarily interleaved content comprehensively.

This work uses LMMs (\ie, \bingname~\citep{bingchat}) for evaluating the open-domain interleaved content, which addresses the above-mentioned issue by accepting multiple image-text pairs and allowing for open-ended evaluation.

\noindent\textbf{Multi-Modal Agents.}
The method part of this paper is related to multi-modal agent studies~\citep{gupta2023visual,suris2023vipergpt,wu2023visual,yang2023mmreact,shen2023hugginggpt,li2023multimodal}, which chain LLMs with multi-modal tools for new tasks.
For example, Visual ChatGPT~\citep{wu2023visual} shows that allocating various generative models~\citep{rombach2022high,meng2021sdedit,zhang2023adding} with ChatGPT~\citep{chatgpt} can achieve complicated image generation and editing. Differently, our work focuses on a specific challenging task of open-domain interleaved image-text generation.

\section{Method}
\label{sec:method}
This section introduces \modelname and the evaluation pipeline based on LMM. Fig.~\ref{fig:overview} overviews our method. Subsequently, we detail our generation and evaluation pipelines, respectively.

\subsection{Interleaved Content Generation}
\label{sec:inter_gen}
\begin{figure}[t]
    \centering
    \begin{subfigure}[t]{\textwidth}
    \centering
    \includegraphics[width=\textwidth]{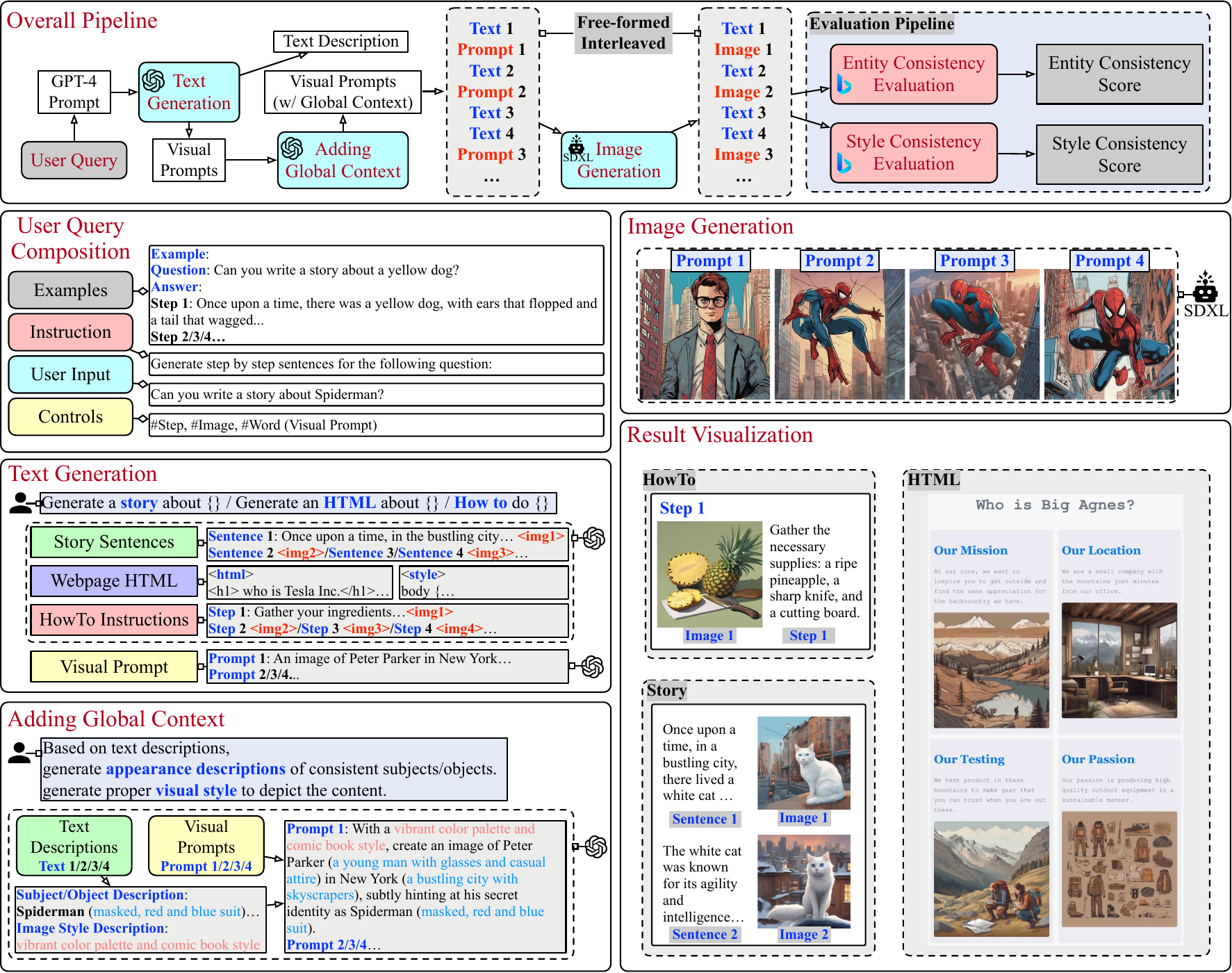}
    \vspace{-5mm}
    \caption{Our interleaved generation framework. The top panel illustrates the overall framework while other panels show details of each procedure.}
    \label{fig:generation}
    \vspace{2mm}
    \end{subfigure}
    \begin{subfigure}[t]{\textwidth}
    \centering
    \includegraphics[width=\textwidth]{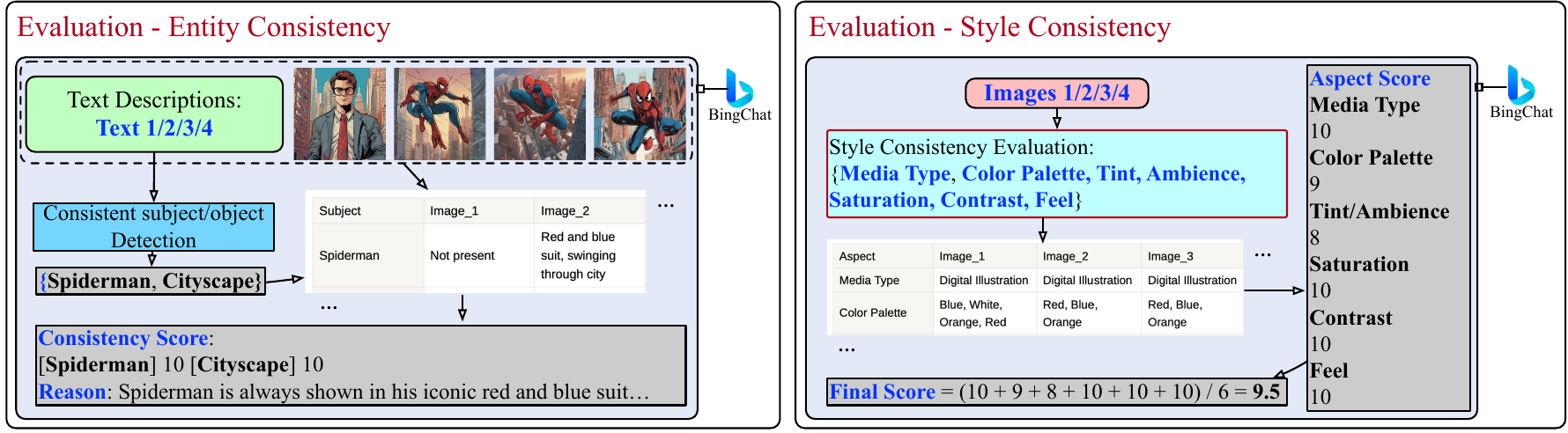}
    \vspace{-5mm}
    \caption{Our LMM-based interleaved evaluation pipeline.}
    \label{fig:evaluation}
    \end{subfigure}
    \vspace{-2mm}
    \caption{Overviews of the proposed interleaved generation framework: (a) and LMM-based evaluation pipeline (b).
    }
    \vspace{-2mm}
    \label{fig:overview}
\end{figure}

We achieve open-domain interleaved generation based on GPT-4~\citep{openai2023gpt4} and SDXL~\citep{podell2023sdxl}. The top panel of Fig.~\ref{fig:generation} shows the generation pipeline of our method. Given an arbitrary user query, we initially follow a meticulously designed composition strategy to assemble an input prompt that indicates the content, format, and constraints of the target output. We then feed the input prompt into GPT-4, which generates the textual descriptions, determines the positions to insert images, and formulates the visual prompt for each image. Subsequently, we incorporate global entity and style contexts into the visual prompts to improve the entity and style consistencies of SDXL. Here, the entity context comprises the appearance descriptions of common subjects, while the style context is a unique image style description shared across all visual prompts. 
Finally, SDXL converts visual prompts into real images, thereby creating the interleaved content.

\noindent\textbf{User Query Composition.}
The input prompts to GPT-4 consist of four parts. We first add a few in-context examples to the start of the prompt. Each example shows the desired output corresponding to a specific input query. 
The in-context examples enable GPT-4 to comprehend the expected content more effectively and encourage it to generate content in the format of the in-context examples, facilitating easier automatic extraction of results. 
Subsequently, we concatenate the generation instruction with the user input to form the prompt. In this case, the instruction tells GPT-4 the desired output type, while the user input specifies the detailed content. 
Finally, we add a few control sentences to the prompt to control the number of image placeholders, story sentences, instruction steps, and \texttt{<div>}s in HTML, \textit{etc}.

\noindent\textbf{Text Generation.}
The first step of \modelname is to generate text. 
By feeding the assembled prompt discussed in the previous part, we enable GPT-4 to produce all text descriptions and image placeholders, indicating the position of each image. 
For example, as shown in the text generation panel of Fig.~\ref{fig:generation}, when generating stories and how-to instructions, GPT-4 is prompted to generate story sentences and instructional steps, respectively, where image tags \texttt{<img\{i\}>} is also included in the generated text. Each image tag indicates the position of the corresponding image, forming an initial interleaved structure. When generating HTML code, the position of each image is determined by the placement of the \texttt{<img>} environment, where the generated CSS code can further tune the size, position, and alignment of each image. 
Next, we prompt GPT-4 to generate visual prompts from text descriptions. In this step, the input prompt also follows the composition strategy introduced earlier, incorporating all story sentences or instructional steps into the user input part. This approach allows GPT-4 to capture the context of the whole story or how-to instructions when generating the visual prompt for each image. 

\noindent\textbf{Adding Global Context.}
To improve the entity and style consistencies of images within the interleaved content, we introduce global entity and style context into the visual prompts before feeding it into SDXL to generate images.
For the global entity context, we add a short appearance description of each common subject to the visual prompts, where GPT-4 is used to extract common subjects from text content, generate appearance descriptions, and rewrite visual prompts. 
To improve the style consistency of images, we prompt GPT-4 to determine a proper visual style to depict the interleaved content, based on the generated text descriptions. For example, GPT-4 indicates that a vibrant color palette and comic book style are best to illustrate superhero stories. Then a short image style description is added to the beginning of each visual prompt to control the artistic style of images generated by SDXL. 
Fig.~\ref{fig:generation} shows the process of adding global context into visual prompts. 
The visual prompts equipped with the global context are then converted into images by SDXL, resulting in the interleaved content.

\begin{figure}[t]
    \centering
    \includegraphics[width=.95\textwidth]{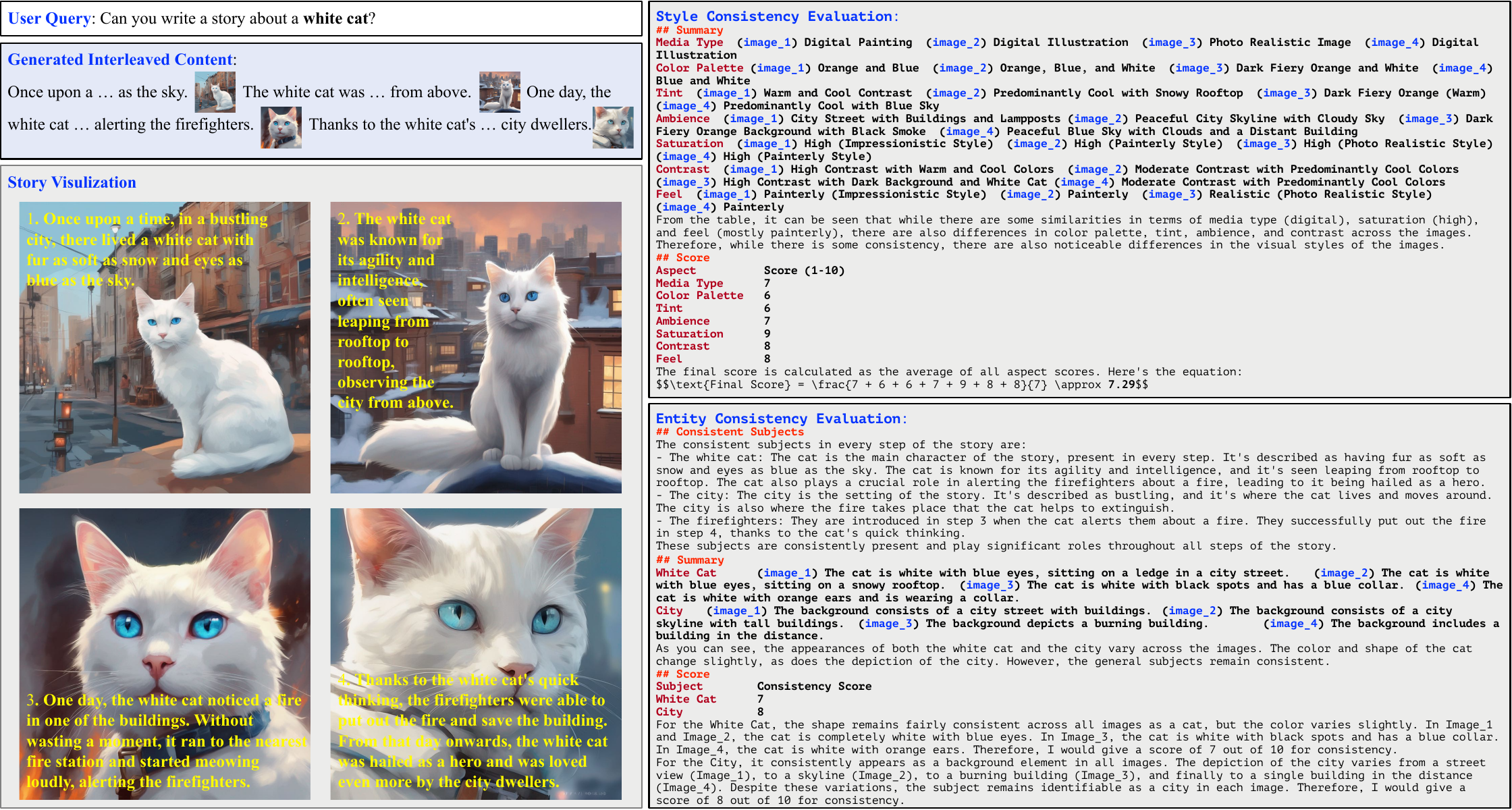}
    \includegraphics[width=.95\textwidth]{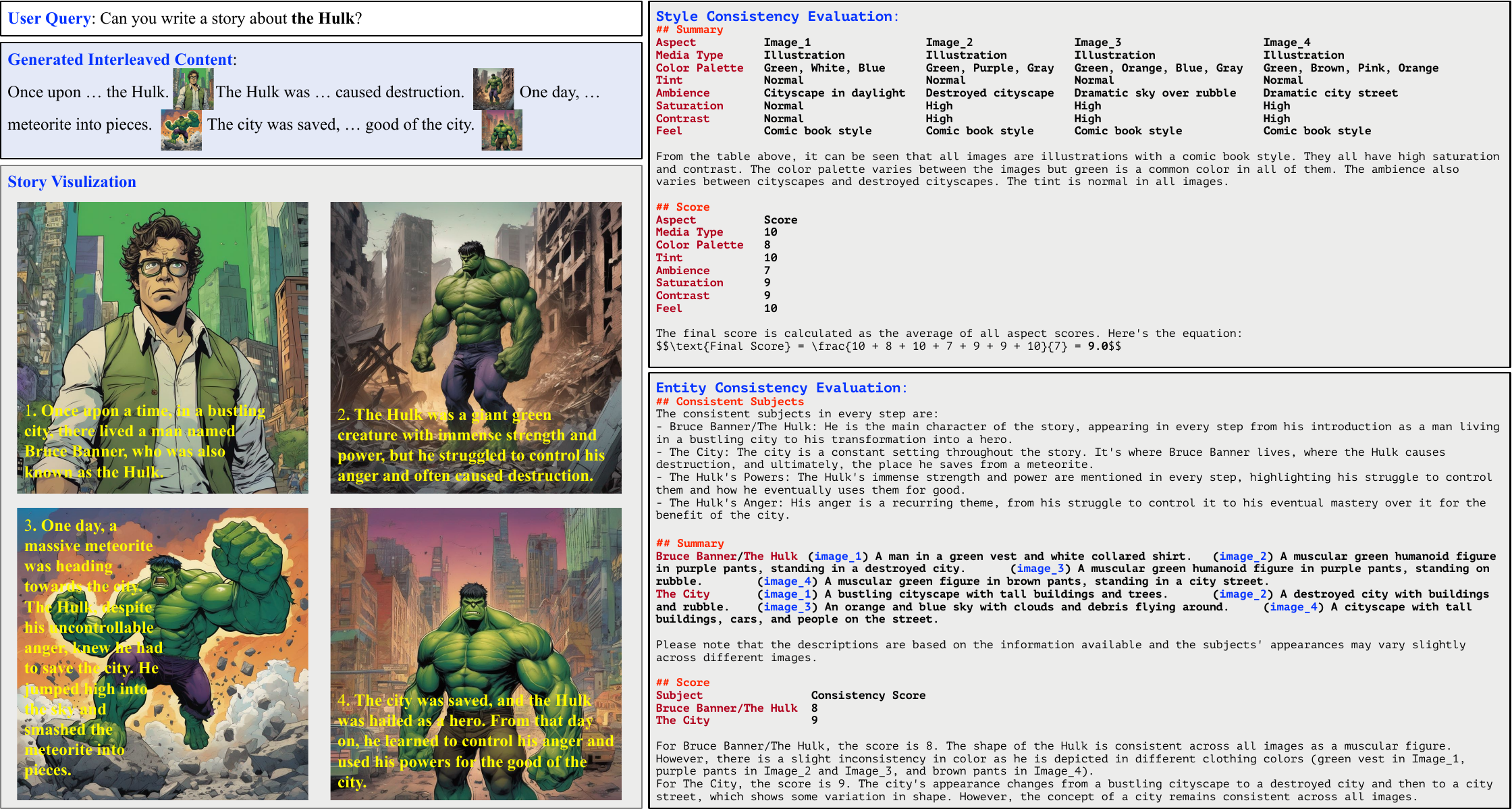}
    \caption{Interleaved visual-language generation results of \modelname on story generation. We visualize the generated interleaved content on the left and the corresponding LMM-Evaluation results on the right. Please zoom in on the screen to see details.}
    \label{fig:main_story}
\end{figure}

\begin{figure}[t]
    \centering
    \vspace{-12mm}
    \includegraphics[width=.95\textwidth]{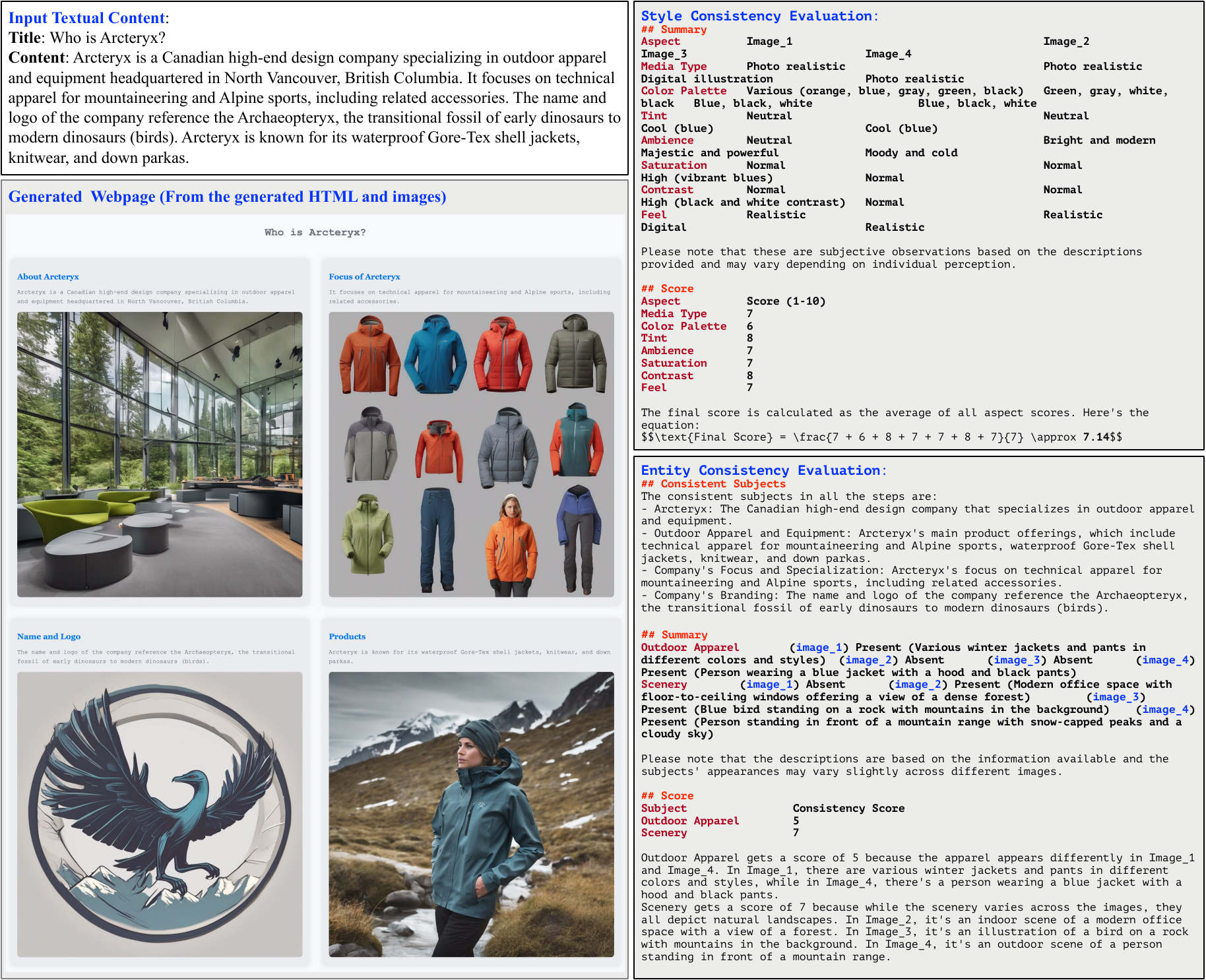}
    \includegraphics[width=.95\textwidth]{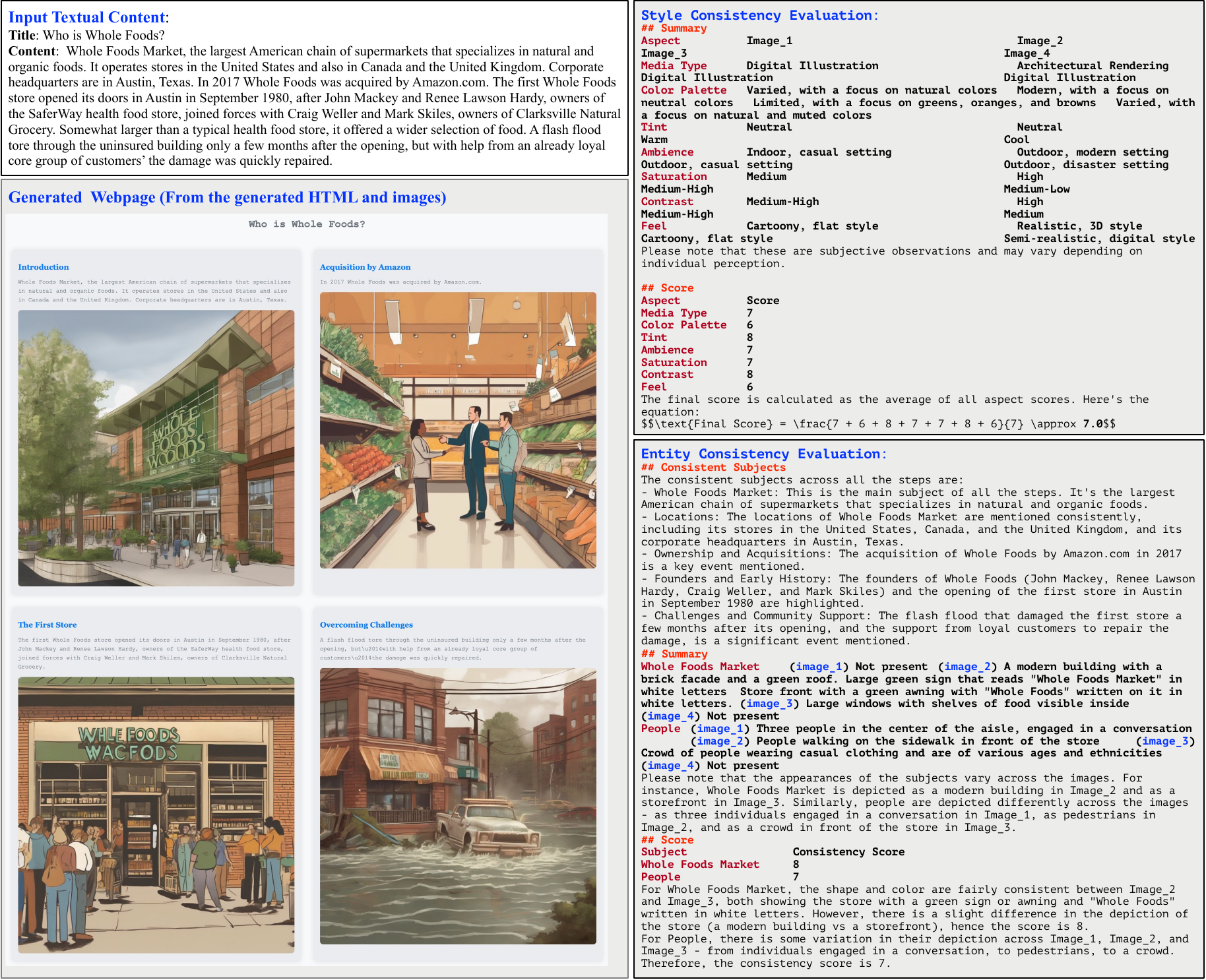}
    \vspace{-2mm}
    \caption{Interleaved visual-language generation results of \modelname on webpage generation. We visualize the generated interleaved content on the left and the corresponding LMM-Evaluation results on the right. Please zoom in on the screen to see details.}
    \label{fig:main_html}
    \vspace{-8mm}
\end{figure}

\begin{figure}[t]
    \centering
    \includegraphics[width=.95\textwidth]{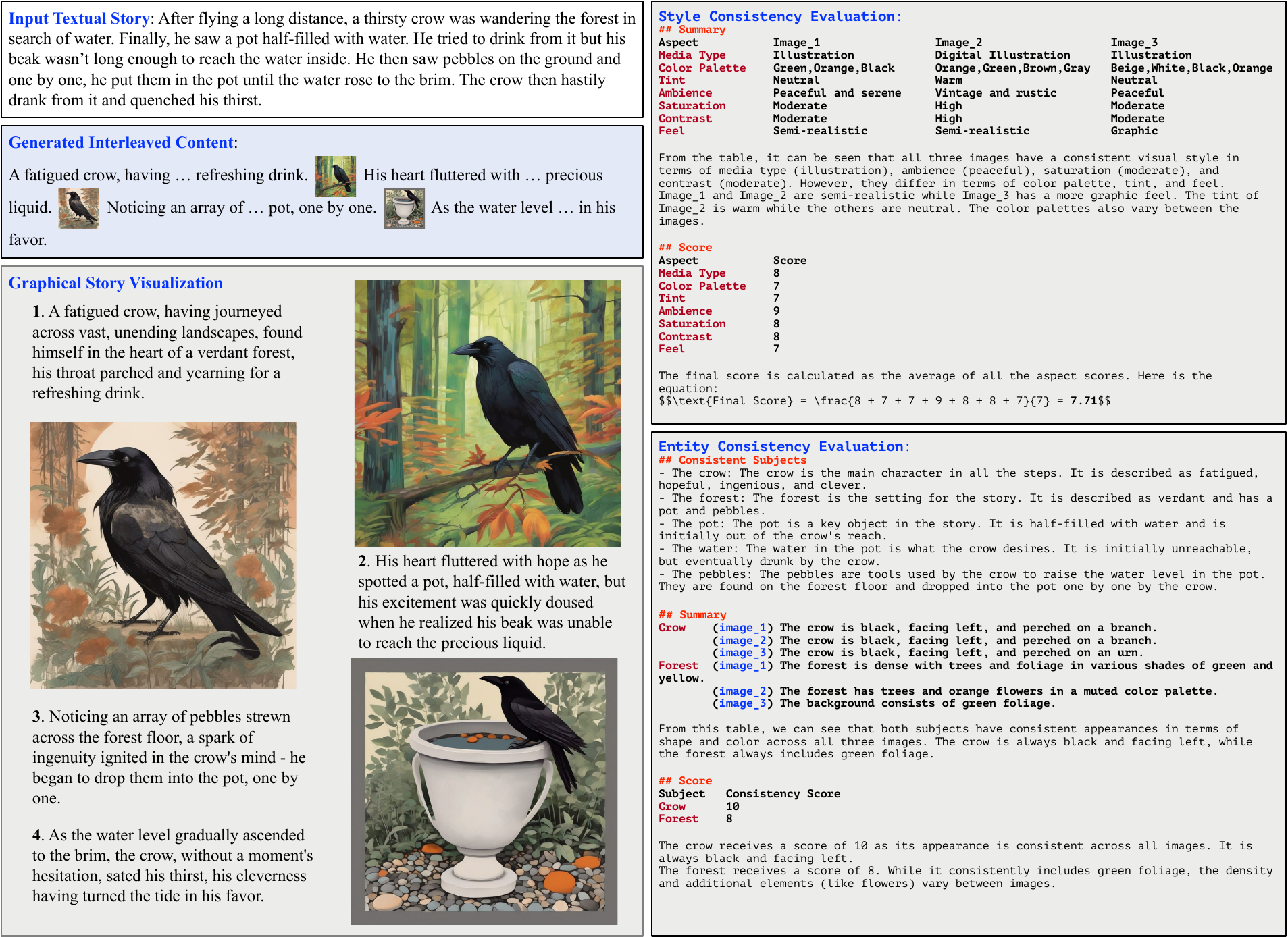}
    \includegraphics[width=.95\textwidth]{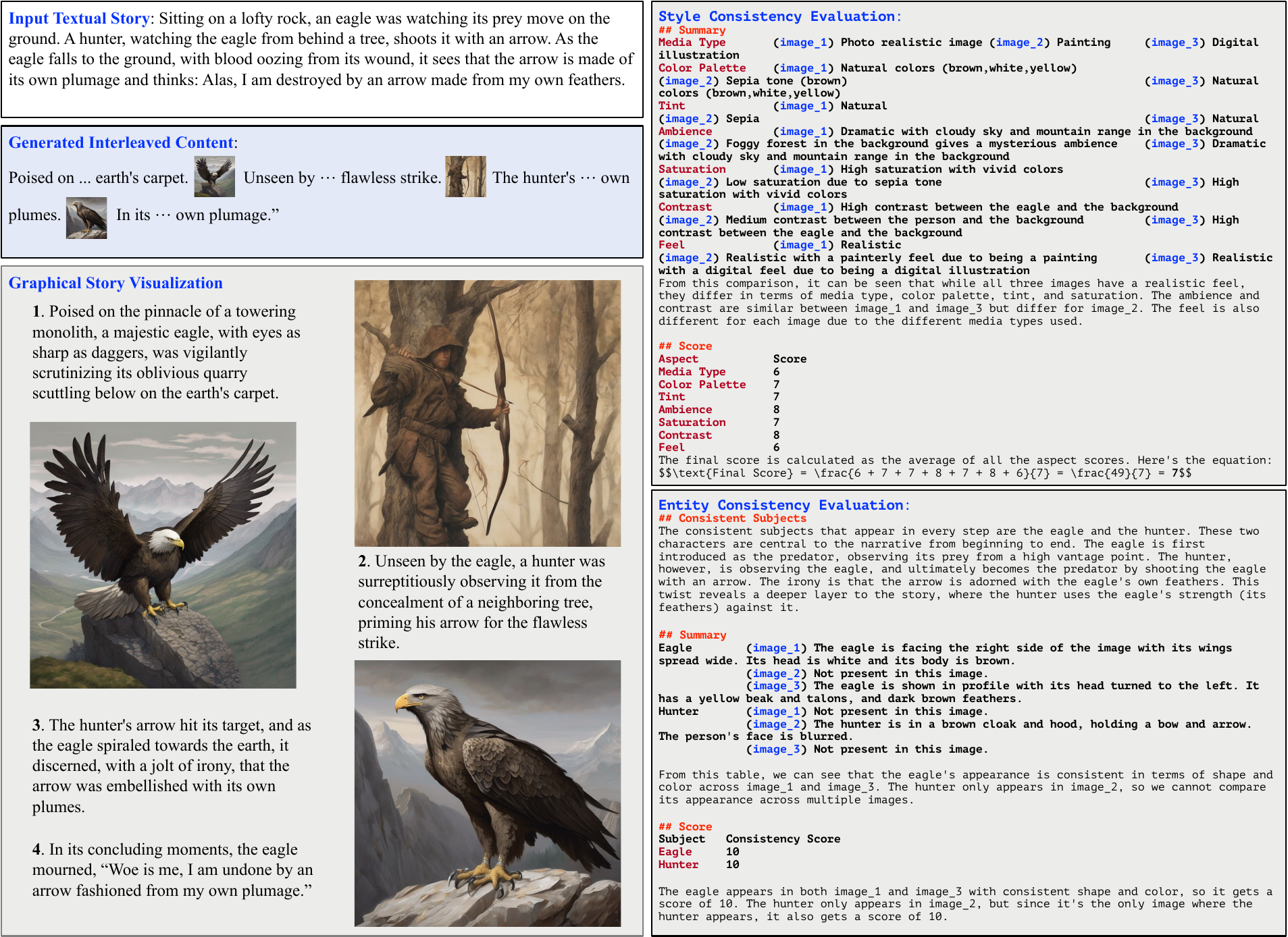}
    \caption{Interleaved visual-language generation results of \modelname on graphical story rewriting. We visualize the generated interleaved content on the left and the corresponding LMM-Evaluation results on the right. Please zoom in on the screen to see details.}
    \label{fig:main_rewrite}
\end{figure}

\begin{figure}[t]
    \centering
    \includegraphics[width=.95\textwidth]{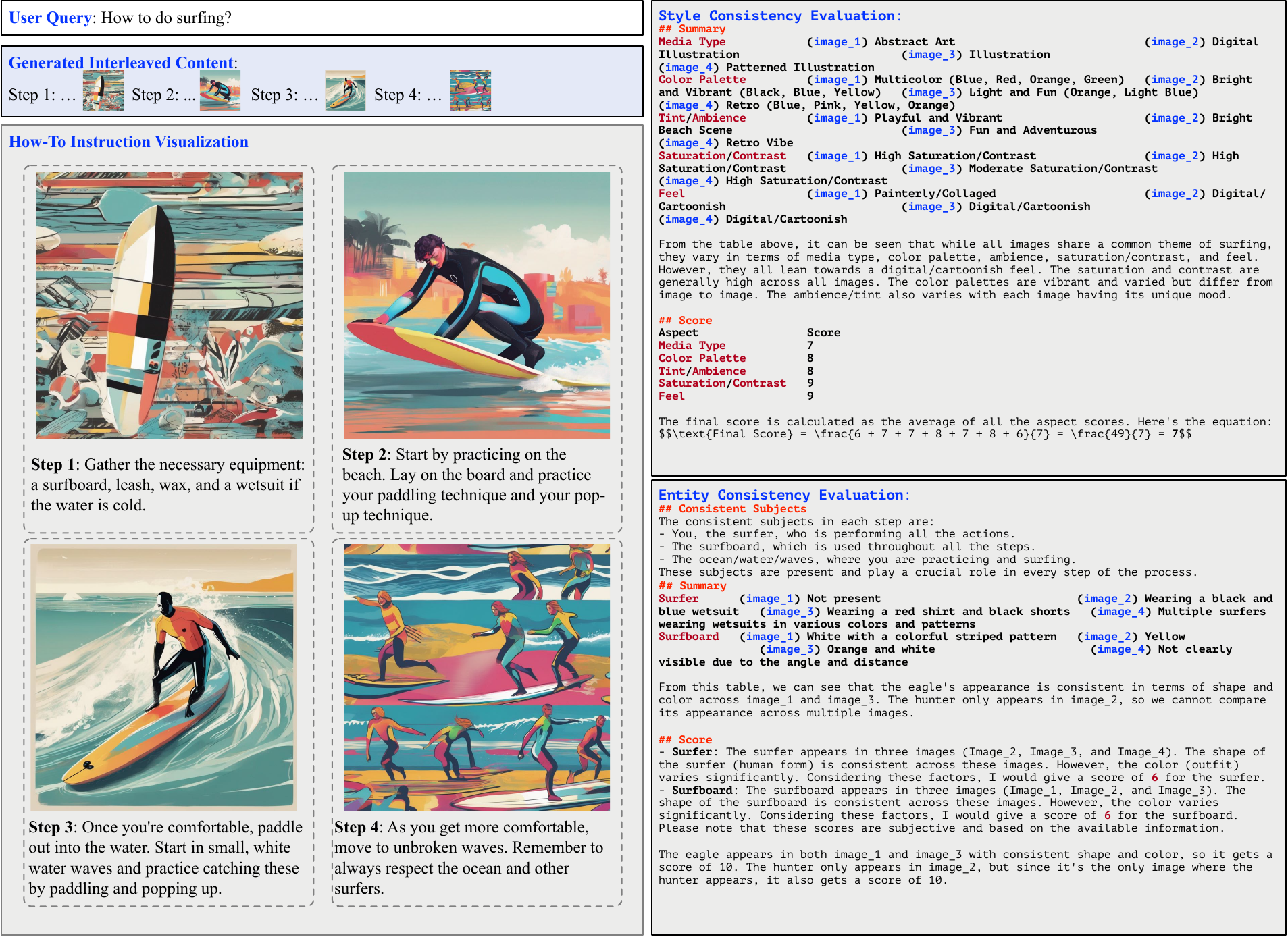}
    \includegraphics[width=.95\textwidth]{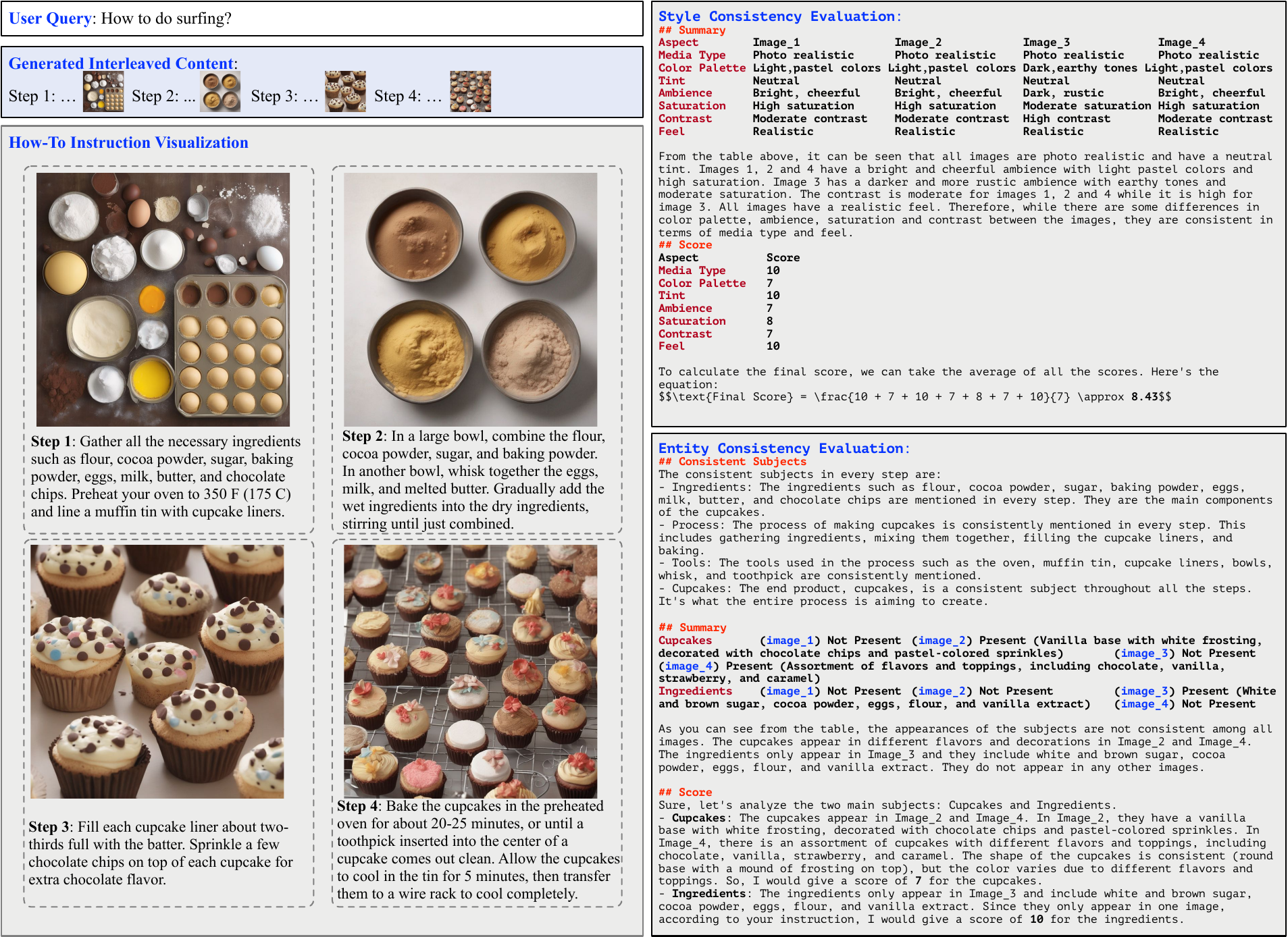}
    \caption{Interleaved visual-language generation results of \modelname on how-to instruction generation. We visualize the generated interleaved content on the left and the corresponding LMM-Evaluation results on the right. Please zoom in on the screen to see details.}
    \label{fig:main_howto}
\end{figure}

\begin{figure}[t]
    \centering
    \includegraphics[width=\textwidth]{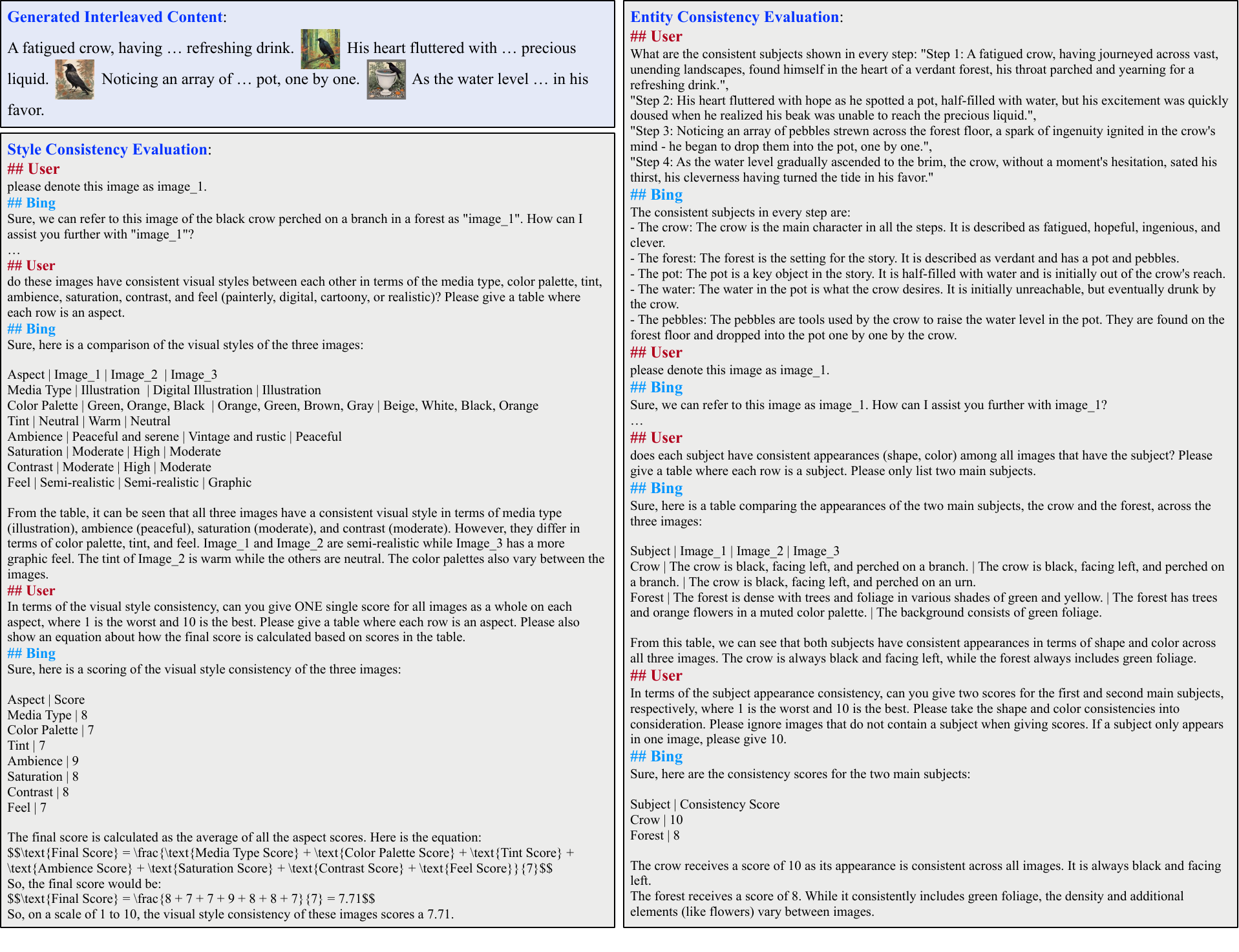}
    \vspace{-8mm}
    \caption{The entity and style consistency evaluation output of \bingname. \bingname can effectively generate rational responses, understand the meaning of pre-defined evaluation aspects, and give scores based on evidence.}
    \vspace{-4mm}
    \label{fig:lmm_eval}
\end{figure}

\begin{figure}[t]
    \centering
    \includegraphics[width=\textwidth]{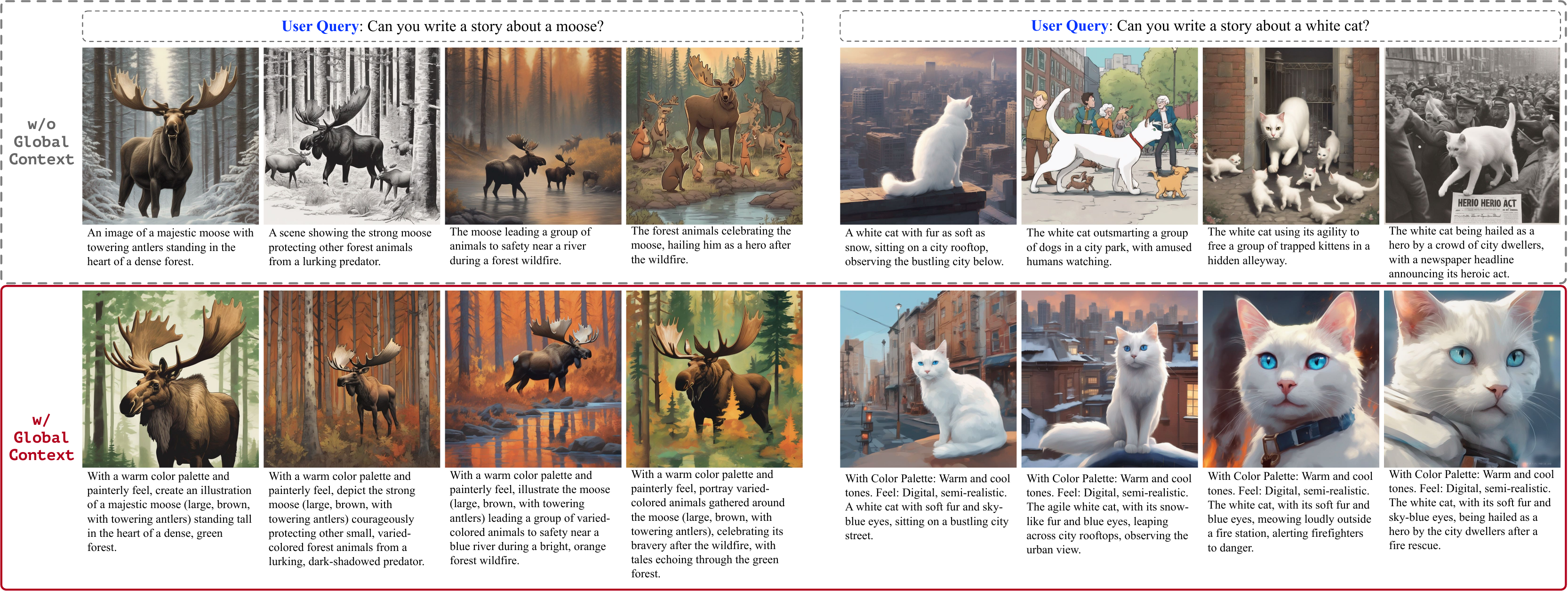}
    \caption{Qualitative image consistency comparison with and without adding global context. We show the T2I prompts used for image generation below the corresponding images.}
    \vspace{-4mm}
    \label{fig:global_context_comp}
\end{figure}

\begin{figure}[t]
    \centering
    \includegraphics[width=\textwidth]{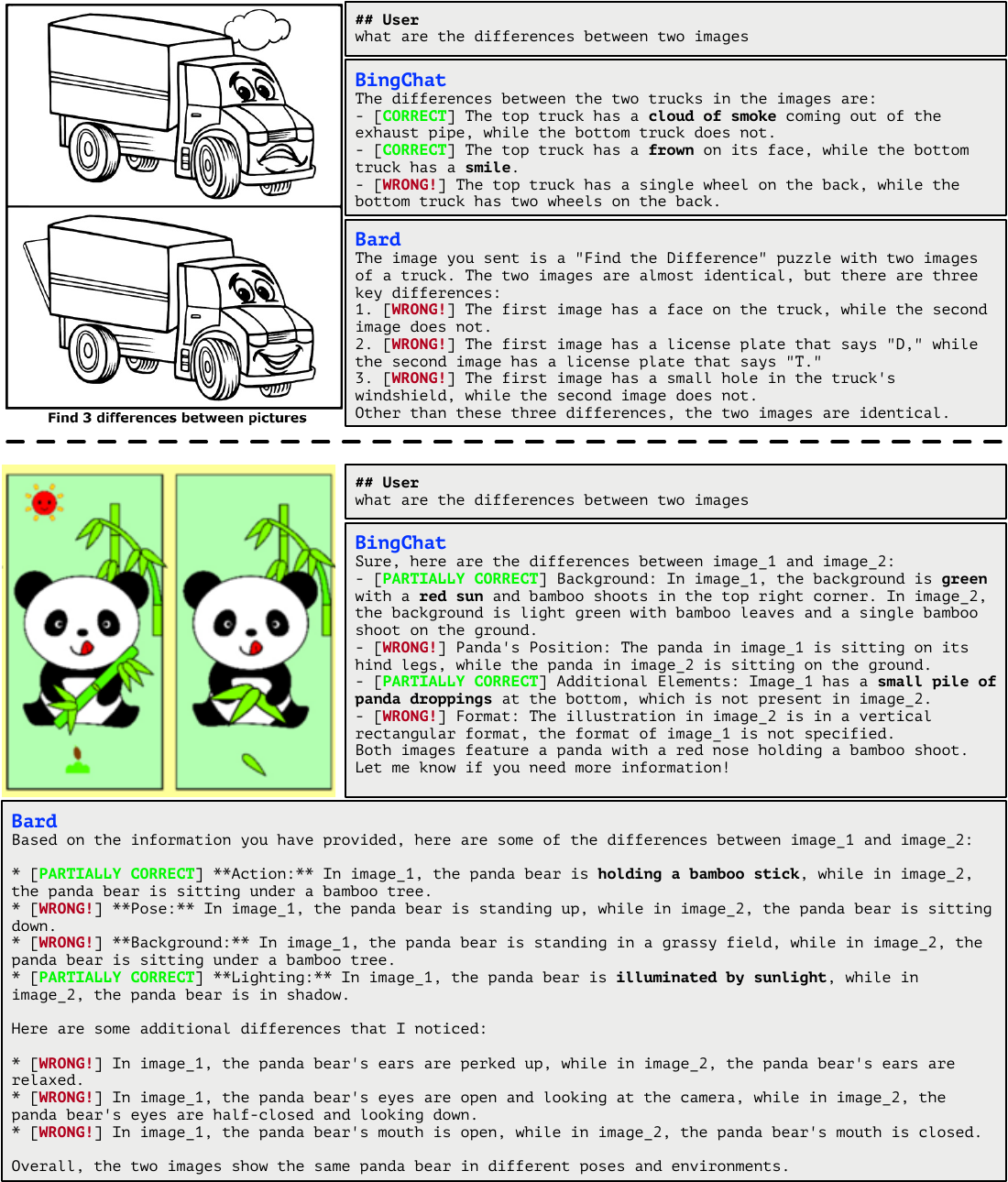}
    \caption{Spot Difference comparison between \bingname and Bard. We empirically observe that \bingname can read the content of images more accurately. Therefore, we use \bingname in our LMM-Evaluation.}
    \vspace{-4mm}
    \label{fig:spot_diff}
\end{figure}

\begin{figure}[t]
    \centering
    \begin{minipage}{.66\textwidth}
    \captionof{table}{The mean and variance of the \bingname evaluation on the benchmark dataset. Adding global context improves the averaged consistencies and lowers the variances.}
    \vspace{-2mm}
    \label{tab:objective}
    \resizebox{\textwidth}{!}{
    \begin{tabular}{l|c|c|c|c}
        \toprule
        \multirow{2}{*}{Model} & \multicolumn{2}{c|}{Entity Consistency} & \multicolumn{2}{c}{Style Consistency} \\ 
        \cline{2-5}
        & \texttt{mean}$\uparrow$ & \texttt{variance}$\downarrow$ & \texttt{mean}$\uparrow$ & \texttt{variance}$\downarrow$ \\
        \midrule
        Ours w/o Global Context & $7.84$ & $1.22$ & $8.00$ & $0.88$ \\
        Ours w/ Global Context & $\mathbf{8.40}$ & $\mathbf{0.77}$ & $\mathbf{8.22}$ & $\mathbf{0.79}$ \\
        \bottomrule
    \end{tabular}
    }
    \captionof{table}{The user preference comparison between \modelname with and without adding global context regarding entity and style consistencies.}
    \label{tab:user_study}
    \resizebox{\textwidth}{!}{
    \begin{tabular}{l|c|c}
        \toprule
        \multirow{2}{*}{Model} & \multicolumn{2}{c}{User Preference (\%)\ $\uparrow$}\\ 
        \cline{2-3}
        &  \texttt{Entity Consistency} & \texttt{Style Consistency}  \\
        \midrule
        Ours w/o Global Context & $35.33$ & $43.35$ \\
        Ours w/ Global Context & $\mathbf{64.67}$ & $\mathbf{56.65}$ \\
        \bottomrule
    \end{tabular}
    }
    \end{minipage}
    \hspace{4mm}
    \begin{minipage}{.29\textwidth}
    \centering
    \includegraphics[width=\textwidth]{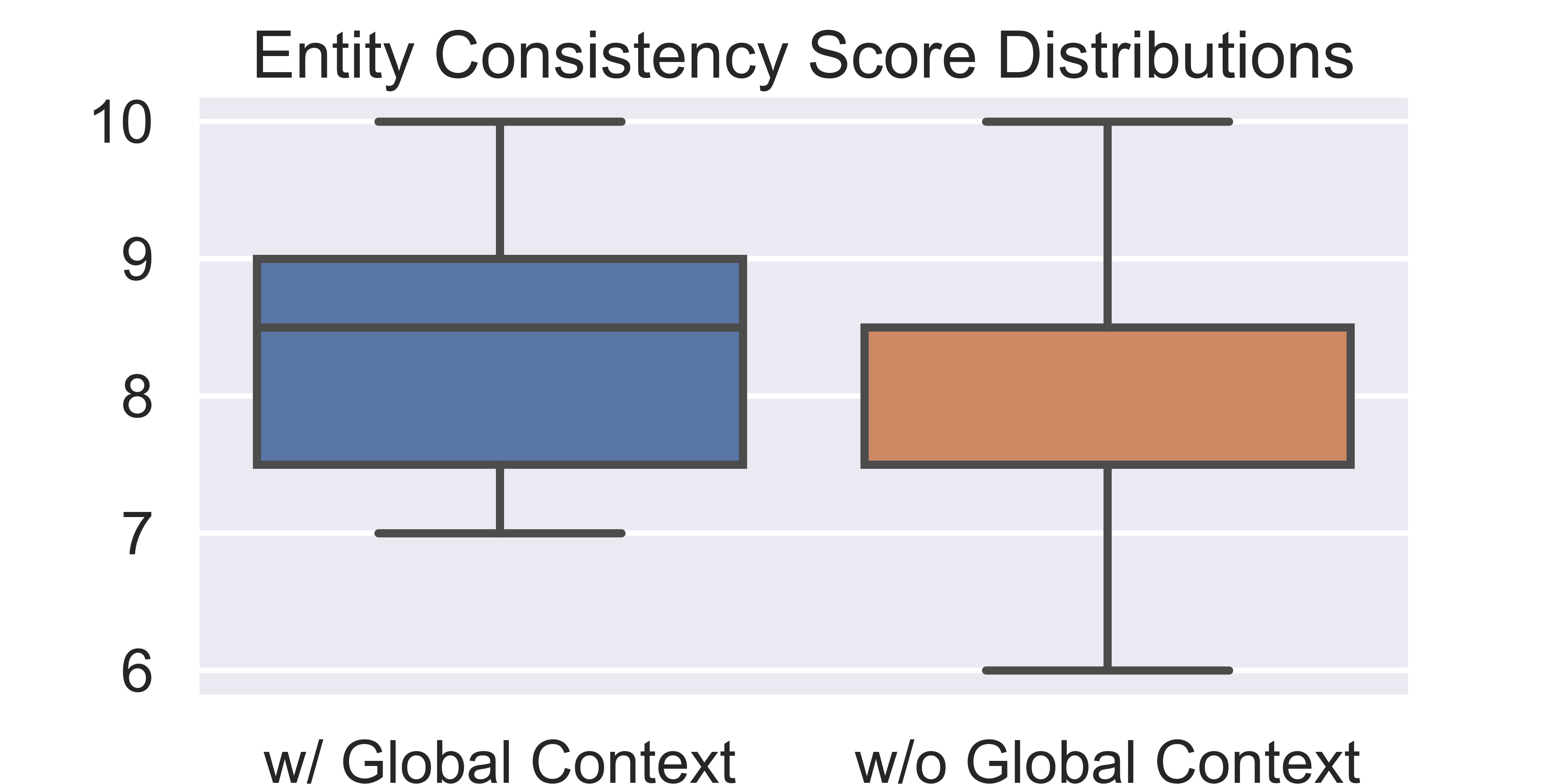}
    \vspace{1mm}
    \includegraphics[width=\textwidth]{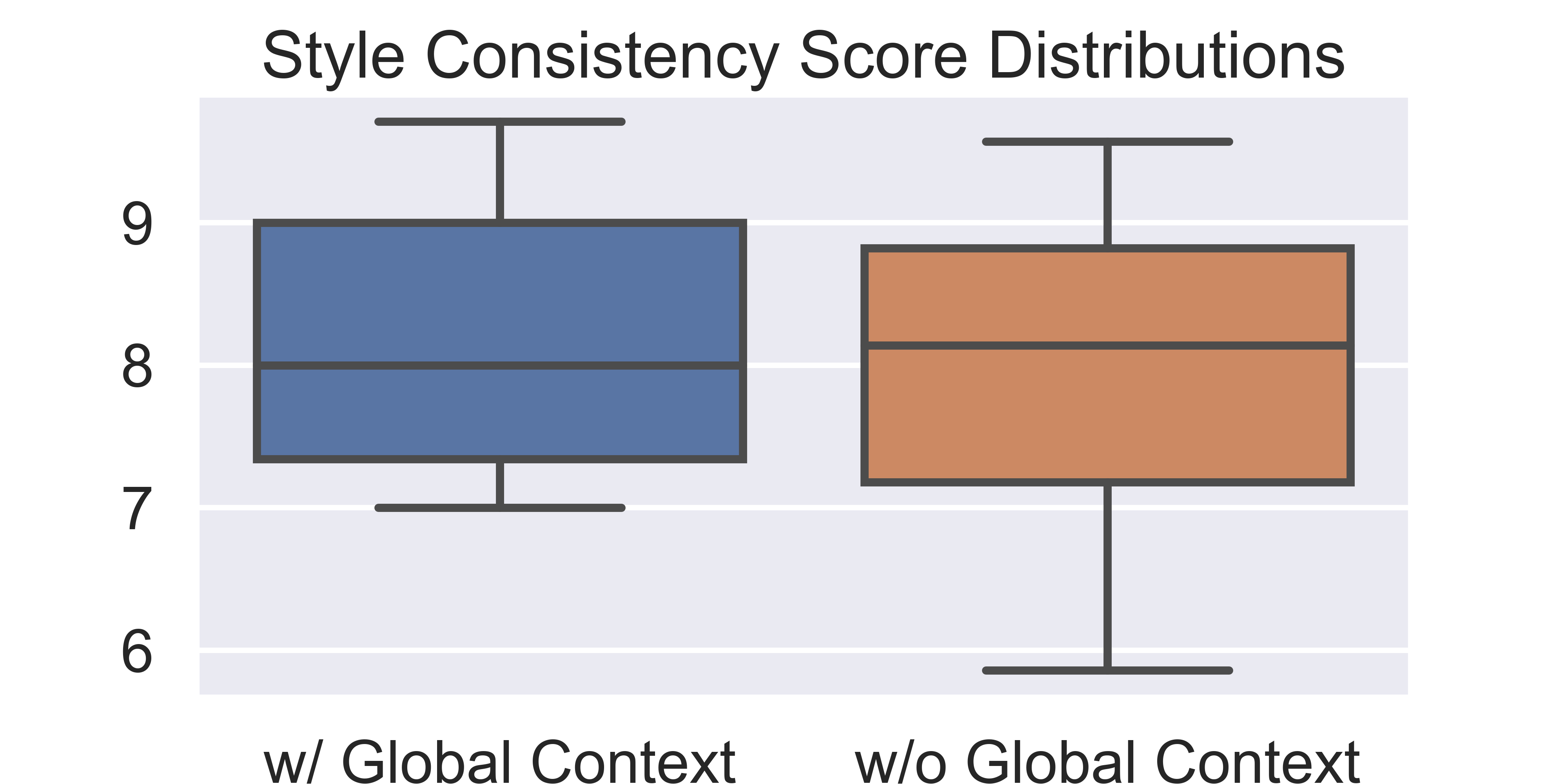}
    \vspace{-5mm}
    \captionof{figure}{The distribution comparison of \bingname evaluation scores.}
    \label{fig:dist}
\end{minipage}
\end{figure}

\subsection{LMM-Based Evaluation}
\label{sec:inter_eval}
We use \bingname~\citep{bingchat} to evaluate the quality of interleaved content based on the chain-of-thought approach, with a focus on the entity and style consistencies. 

\noindent\textbf{Entity Consistency Evaluation.}
To evaluate the entity consistency of images within the interleaved content. We first prompt~\bingname to detect two main common subjects from the generated text descriptions. For example, in Fig.~\ref{fig:evaluation}, the Spiderman and the background cityscape are the main subjects in images, which should have consistent entities and appearances between images. Subsequently, 
we input all images into~\bingname , assigning a unique index \texttt{image\_\{i\}} to each image.
We then have~\bingname summarize the appearance of each common subject in the images and assign a score for the entity consistency based on the appearances of common subjects. 

\noindent\textbf{Style Consistency Evaluation}
We evaluate the style consistency of images based on seven visual factors: media type, color palette, tint, ambiance, saturation, contrast, and overall feel of images. The media type and feel recognize the global feeling of an image such as a realistic photo, painting, digital illustration, or cartoon. While the other evaluation factors detect more subtle visual style distinctions within images. 
Similar to the entity evaluation, we use~\bingname to summarize the visual style of each image based on the pre-defined visual factors and score the style consistency for each factor. The final score is calculated as the mean value of the consistency score on all visual factors.
\section{Experiments}
This section introduces our experimental settings, results, and analysis. 

\subsection{Experiment Settings}
\noindent\textbf{Implementation Details.}
We utilize the text-only GPT-4 API of July 2023. For T2I generation, we use the text-to-image pipeline of the SDXL v$1.0$ model~\footnote{\url{https://huggingface.co/stabilityai/stable-diffusion-xl-base-1.0}}, with all hyper-parameters set to their defaults.
For LMM-Evaluation, we apply the precise mode of~\bingname to both entity and style consistency evaluations.
We develop a Python code to connect all the above prompting procedures, which automatically extracts the desired content from the previous model's output and forms the input for the next procedure.

\noindent\textbf{Compared Model Variants.}
To demonstrate the effectiveness of the global context, we compare the performance of the proposed baseline and a simplified variant without the global context.

\noindent\textbf{Evaluation Set.}
We collect a benchmark dataset of thirty problems, which covers four interleaved generation tasks: graphical storytelling, visual how-to instruction generation, text-to-graphic story rewriting, and webpage generation. The how-to instruction generation task has ten problems, which include the requests to generate the recipes for foods and the tutorials for sports and actions. The webpage generation task has five questions about converting a company introduction into a webpage or poster. 
The ten storytelling and five graphical story rewriting problems ask the model to generate animal- or superhero-related stories. Please refer to the supplementary material for problems of the constructed benchmark.

\subsection{Experiment Results}
\noindent\textbf{Selection of LMMs.} 
\bingname~\citep{bingchat} and Bard~\citep{bard} are two publically available LMMs that have the image reading ability. The accuracy of LMM's image understanding plays a pivotal role in our approach to evaluating generated interleaved content.
As shown in Fig.~\ref{fig:spot_diff}, we conduct a spot-the-difference test to understand the image understanding ability of the two LMMs. The test requires the ability to compare two images in detail, which is essential for effective evaluation. We empirically observe that \bingname (2023-10-09) performs better in the spot-the-difference task than Bard (2023-10-09), and thus opt for \bingname for the LMM evaluation. For example, in the top example of Fig.~\ref{fig:spot_diff}, \bingname successfully finds two out of three differences between two images while Bard fails. In the bottom example, \bingname spots the red sun in the left image, but Bard only gives an ambiguous description of the sunlight difference. Furthermore, all the additional differences reported by Bard are out of the content of given images, showing a more noticeable hallucination issue than \bingname.

\noindent\textbf{Qualitative Results.}
Fig.~\ref{fig:main_story},~\ref{fig:main_html},~\ref{fig:main_rewrite},~\ref{fig:main_howto} show the qualitative results of the \modelname. We observe that \modelname generates coherent interleaved image-text sequences based on arbitrary input queries. For example, in Fig.~\ref{fig:main_story}, \modelname generates coherent and interesting multi-modal stories about a white cat and the Hulk. All images have consistent entities and styles while being visually appealing. Fig.~\ref{fig:main_html} shows \modelname generating commercial webpages based on introductions of ``Arcteryx'' and ``Whole Foods''. \modelname naturally splits text introductions into paragraphs and accurately adds paragraph headlines. The generated HTML and CSS files arrange the interleaved content in an intuitive and attractive way. Fig.~\ref{fig:main_rewrite} shows \modelname converting a textual story into multi-modal story and Fig.~\ref{fig:main_howto} shows generating visual how-to instruction. The story sentences in the graphical story become more vivid compared with the input textual story while visually appealing and coherent images are naturally inserted into arbitrary text locations. The generated visual how-to instructions are informative and the illustration images are helpful in understanding terminologies of surfing.

More importantly, Fig.~\ref{fig:lmm_eval} shows that the \bingname is effective in evaluating entity and style consistencies, even for the complicated interleaved image-text sequences. For example, in evaluating entity consistency, \bingname accurately grounds common subjects extracted from text descriptions to the corresponding subjects in images. We observe that \bingname are capable of generating rationale text, understanding the meaning of and giving consistency scores on pre-defined sub-categories, as well as calculating the final score based on a clear mathematical equation. 

\noindent\textbf{Effectiveness of Adding Global Context.} To qualitatively demonstrate the effectiveness of adding global context, we compare the generated images in the interleaved content in Fig.~\ref{fig:global_context_comp}. We observe that adding global context can effectively improve the style and entity consistencies. For example, in the left example of a story of moose, the second image has a black-white color palette while the fourth image has a cartoon feel, which violets the overall vibrant color palette and realistic illustration feel in other images. However, the bottom images generated with the global context show much better style consistency with a consistent illustration feel and vivid color palette. On the right-side example, adding global context improves the entity consistency of the cat as the main character of the story as well as the image's visual style.

\noindent\textbf{Quantitative Results.}
We then evaluate \modelname and its variants using the effective \bingname evaluation. We compute the entity and style consistency scores on the interleaved content generated from problems in the benchmark set. Table~\ref{tab:objective} shows the mean and variance values of the entity and style consistency scores over the benchmark set while the boxplots in Fig.~\ref{fig:dist} show the comparison of the score distributions. Both \modelname and its variant show good entity and style consistencies. After adding the global context, we observe that the mean values of both entity and style consistency scores increased, indicating improved consistencies. On the other hand, the global context also reduced the variances of both entity and style consistency scores. demonstrating that adding global context can improve the robustness of \modelname in terms of content and style consistencies. 

\noindent\textbf{User Study Results.} In addition to the quantitative evaluation based on \bingname, we conduct a user study to subjectively evaluate the performance of \modelname and its variants. The user study contains two sections, where the first section contains $24$ questions, asking users to evaluate the entity consistency of generated image sequences in the interleaved content. The second section is to evaluate the style consistency, which have $29$ questions. In each question of the user study, we show two image sequences generated by \modelname and the variant without adding the global context given the same user input query. We let the user to choose the one with better entity or style consistencies in the first and second sections, respectively. We collect overall $167$ and $203$ votes for the entity and style consistencies, respectively. Table~\ref{tab:user_study} shows the user preference percentages regarding the entity and style consistencies. \modelname outperforms its variant without using the global context in both entity and style consistencies, which, on the one hand, demonstrates that adding global context is effective in improve the image consistency in the interleaved content generation. On the other hand, it also verifies that the proposed LMM-based evaluation can produce evaluation results aligned with the human preferences.

\begin{table}[t]
    \centering
    \captionof{table}{The correlation analysis between the human and \bingname evaluations. $p$ denotes the p-value.}
    \label{tab:correlation}
    \setlength\tabcolsep{20pt}
    \resizebox{1\linewidth}{!}{
    \begin{tabular}{l|c|c}
        \toprule
        Correlation Index & Entity Score & Style Score \\ 
        \midrule
        Kendall’s Tau$\uparrow$ $\in\left(-1, 1\right)$ & $0.87$ ($p=0.0008$) & $0.58$ ($p=0.0196$)  \\
        Spearman's Correlation$\uparrow$ $\in\left(-1, 1\right)$ & $0.78$ ($p=0.0080$) & $0.95$ ($p=0.0000$) \\
        \bottomrule
    \end{tabular}
    }
\end{table}

\subsection{Evaluation Pipeline Analysis}
To ensure that the LLM-evaluation pipeline can produce the entity and style consistency scores in line with the human perception. We conduct an analysis to study the score correlation between LLM-evaluation and the human rating. We collect ten image pairs from our generated interleaved content, where each pair comes from the same story. First, we use LLM-evaluation to give both entity and style consistency scores on each image pair. Next, we conduct a user study to let users rate the entity and style consistencies of each image pair. Considering that humans may have difficulty giving an absolute score of consistency, in the user study, we randomly select two image pairs from the evaluation set and show them side-by-side to users, letting users choose one pair that has better entity or style consistency. 
Finally, we calculate the consistency scores of each image pair based on the number of user preferences. 

To determine whether the proposed LLM-evaluation aligns well with human perception, we compute Kendall's Tau~\citep{kendall1938new} and Spearman's Correlation~\citep{kendall1973advanced} indexes, which measure the similarity between the rankings from the human and LLM-evaluation scores. Table~\ref{tab:correlation} shows the correlation scores. For both entity and style consistencies, we achieve Kendall's Tau and Spearman Rank-Order close to $1$ with low p-values, strongly indicating that our LLM-evaluation method has a good alignment with the human evaluation in both measuring entity and style consistencies of the interleaved content.

\section{Conclusion}
In this paper, we fill the gap of open-domain interleaved image-text generation by introducing a baseline interleaved generation method \modelname based on GPT-4 and SDXL, an evaluation pipeline based on \bingname, and a benchmark dataset to compare different approaches. Experimental results on the constructed benchmark dataset show that \modelname has a strong ability to generate arbitrarily interleaved image-text sequences for addressing open-domain user queries. A comprehensive analysis based on the user study demonstrates that the evaluation method based on \bingname can effectively capture the styles and entities within images, thereby reliably evaluating generated interleaved multimodal contents.

\noindent\textbf{Ethics Statement.} In this research on the open-domain interleaved image-text generation, we uphold a commitment to ethical conduct guided by principles prioritizing human welfare, privacy, fairness, accountability, and transparency. Our model is built on top of GPT-4, BingChat, and Stable Diffusion XL, where all the pre-trained models are publicly available and we believe their owners are aware of addressing the potential ethics issues. Our method does not bring new ethical issues. On the data side, we do not include human-related problems in our constructed interleaved benchmark dataset to avoid potential privacy and ethics issues.

\noindent\textbf{Reproducibility Statement.} All the pre-trained models we used are publicly available, where we indicate the model version, settings, and hyper-parameters we used in our experiments. We believe our method and the evaluation pipeline are reproducible.

\appendix
\section{Evaluation Problems}
Table~\ref{tab:sup_tab1},~\ref{tab:sup_tab2}, and~\ref{tab:sup_tab3} show the complete list of problems in our constructed evaluation set.
\begin{table}[ht]
    \centering
    \caption{Benchmark problem list of how-to instruction generation and story telling.}
    \begin{tabular}{p{4cm}|p{8cm}}
    \toprule
    Task & Problem \\
    \midrule
       \multirow{10}{*}{How-To Instructions}  &  How to make monkey brains cocktail? \\
& How to make Korean banana milk?\\
& How to cook charcoal-grilled ribeye steak?\\
& How to make classic chocolate cupcakes?\\
& How to make ice-cream cake pops?\\
& How to make perfect gooey cinnamon rolls?\\
& How to do surfing?\\
& How to clean a glass bowl?\\
& How to trim a tree?\\
& How to cut a pineapple into pieces?\\
\midrule
\multirow{10}{*}{Story Telling} & Can you write a story about the Spiderman?\\
&Can you write a story about the Iron Man?\\
&Can you write a story about the Superman?\\
&Can you write a story about the Batman?\\
&Can you write a story about the Hulk?\\
&Can you write a story about a white cat?\\
&Can you write a story about a giraffe?\\
&Can you write a story about a penguin?\\
&Can you write a story about a polar bear?\\
&Can you write a story about a moose?\\
\bottomrule
    \end{tabular}
    \label{tab:sup_tab1}
\end{table}

\begin{table}[ht]
    \centering
    \caption{Benchmark problem list of graphical story rewriting.}
    \begin{tabular}{p{12cm}}
    \toprule
    Textual Story \\
    \midrule
After flying a long distance, a thirsty crow was wandering the forest in search of water. Finally, he saw a pot half-filled with water. He tried to drink from it but his beak wasn’t long enough to reach the water inside. He then saw pebbles on the ground and one by one, he put them in the pot until the water rose to the brim. The crow then hastily drank from it and quenched his thirst.\\
\midrule
One day, a camel and her baby were chatting. The baby asked, “Mother, why do we have humps?” The mother replied, “Our humps are for storing water so that we can survive in the desert”. “Oh”, said the child, “and why do we have rounded feet mother?” “Because they are meant to help us walk comfortably in the desert. These legs help us move around in the sand.” “Alright. But why are our eyelashes so long?” “To protect our eyes from the desert dust and sand. They are the protective covers for the eyes”, replied the mother camel. The baby camel thought for a while and said, “So we have humps to store water for desert journeys, rounded hooves to keep us comfortable when we walk in the desert sand, and long eyelashes to protect us from sand and dust during a desert storm. Then what are we doing in a zoo?” The mother was dumbfounded.\\
\midrule
One day, two goats try to cross a weak and narrow bridge across the river. The goats are at either end of the bridge, but neither is ready to make way for the other. They come to the centre of the bridge and begin fighting about who should cross first. As they fight mindlessly, the bridge gives in, taking both the goats down into the river with it.\\
\midrule
Sitting on a lofty rock, an eagle was watching its prey move on the ground. A hunter, watching the eagle from behind a tree, shoots it with an arrow. As the eagle falls to the ground, with blood oozing from its wound, it sees that the arrow is made of its own plumage and thinks: “Alas, I am destroyed by an arrow made from my own feathers”.\\
\midrule
A fox sees a crow carrying a piece of cheese to a tree top. It decides to get the cheese for himself. It goes to the tree and starts praising the crow that it can sing better than a cuckoo. Hearing this, the crow beams with pride, and tries to sing. The piece of cheese falls to the ground as it opens its mouth to sing. The fox picks up the piece and runs away.\\
\bottomrule
    \end{tabular}
    \label{tab:sup_tab2}
\end{table}

\begin{table}[ht]
    \centering
    \caption{Benchmark problem list of webpage generation.}
    \begin{tabular}{p{12cm}}
    \toprule
    Webpage Textual Content \\
    \midrule
Title: Who is Arcteryx? Content: Arcteryx is a Canadian high-end design company specializing in outdoor apparel and equipment headquartered in North Vancouver, British Columbia. It focuses on technical apparel for mountaineering and Alpine sports, including related accessories. The name and logo of the company reference the Archaeopteryx, the transitional fossil of early dinosaurs to modern dinosaurs (birds). Arcteryx is known for its waterproof Gore-Tex shell jackets, knitwear, and down parkas.\\
\midrule
Title: Who is Whole Foods? Content: Whole Foods Market, the largest American chain of supermarkets that specializes in natural and organic foods. It operates stores in the United States and also in Canada and the United Kingdom. Corporate headquarters are in Austin, Texas. In 2017 Whole Foods was acquired by Amazon.com. The first Whole Foods store opened its doors in Austin in September 1980, after John Mackey and Renee Lawson Hardy, owners of the SaferWay health food store, joined forces with Craig Weller and Mark Skiles, owners of Clarksville Natural Grocery. Somewhat larger than a typical health food store, it offered a wider selection of food. A flash flood tore through the uninsured building only a few months after the opening, but—with help from an already loyal core group of customers—the damage was quickly repaired.\\
\midrule
Title: Who is Marvel Universe? Content: The Marvel Universe is a fictional shared universe where the stories in most American comic book titles and other media published by Marvel Comics take place. Super-teams such as the Avengers, the X-Men, the Fantastic Four, the Guardians of the Galaxy, and many Marvel superheroes live in this universe, including characters such as Spider-Man, Captain America, Iron Man, Thor, the Hulk, Ant-Man, the Wasp, Wolverine, Black Panther, Doctor Strange, Daredevil, and Captain Marvel, Blade, Black Widow, Hawkeye, among numerous others. It also contains well-known supervillains such as Doctor Doom, Magneto, Ultron, Thanos, Loki, The Green Goblin, Kang the Conqueror, Red Skull, The Kingpin, Doctor Octopus, Carnage, Apocalypse, Dormammu, Mysterio, Electro, and the Vulture. It also contains antiheroes such as Venom, Namor, Deadpool, Silver Sable, Ghost Rider, The Punisher, and Black Cat.\\
\midrule
Title: Who is big agnes? Content: At our core, we want to inspire you to get outside and find the same appreciation for the backcountry we have. We are a small company with the mountains just minutes from our office. We test product in these mountains to make gear that you can trust when you are out there. Our passion is producing high quality outdoor equipment in a sustainable manner. We outfit all people with the gear needed to camp comfortably, explore the backcountry and have FUN!\\
\midrule
Title: Who is Tesla, Inc.? Content: Tesla, Inc. is an American multinational automotive and clean energy company headquartered in Austin, Texas. Tesla designs and manufactures electric vehicles (cars and trucks), stationary battery energy storage devices from home to grid-scale, solar panels and solar shingles, and related products and services. Its subsidiary Tesla Energy develops and is a major installer of photovoltaic systems in the United States and is one of the largest global suppliers of battery energy storage systems with 6.5 gigawatt-hours (GWh) installed in 2022.\\
\bottomrule
    \end{tabular}
    \label{tab:sup_tab3}
\end{table}
\clearpage

\bibliography{ref}

\begin{thebibliography}{41}
\providecommand{\natexlab}[1]{#1}
\providecommand{\url}[1]{\texttt{#1}}
\expandafter\ifx\csname urlstyle\endcsname\relax
  \providecommand{\doi}[1]{doi: #1}\else
  \providecommand{\doi}{doi: \begingroup \urlstyle{rm}\Url}\fi

\bibitem[Black et~al.(2023)Black, Janner, Du, Kostrikov, and
  Levine]{black2023training}
Kevin Black, Michael Janner, Yilun Du, Ilya Kostrikov, and Sergey Levine.
\newblock Training diffusion models with reinforcement learning.
\newblock \emph{arXiv preprint arXiv:2305.13301}, 2023.

\bibitem[Chiang \& Lee(2023)Chiang and Lee]{chiang2023can}
Cheng-Han Chiang and Hung-yi Lee.
\newblock Can large language models be an alternative to human evaluations?
\newblock \emph{arXiv preprint arXiv:2305.01937}, 2023.

\bibitem[Dong et~al.(2023)Dong, Han, Peng, Qi, Ge, Yang, Zhao, Sun, Zhou, Wei,
  et~al.]{dong2023dreamllm}
Runpei Dong, Chunrui Han, Yuang Peng, Zekun Qi, Zheng Ge, Jinrong Yang, Liang
  Zhao, Jianjian Sun, Hongyu Zhou, Haoran Wei, et~al.
\newblock Dreamllm: Synergistic multimodal comprehension and creation.
\newblock \emph{arXiv preprint arXiv:2309.11499}, 2023.

\bibitem[Fu et~al.(2023)Fu, Ng, Jiang, and Liu]{fu2023gptscore}
Jinlan Fu, See-Kiong Ng, Zhengbao Jiang, and Pengfei Liu.
\newblock Gptscore: Evaluate as you desire.
\newblock \emph{arXiv preprint arXiv:2302.04166}, 2023.

\bibitem[Gadre et~al.(2023)Gadre, Ilharco, Fang, Hayase, Smyrnis, Nguyen,
  Marten, Wortsman, Ghosh, Zhang, et~al.]{gadre2023datacomp}
Samir~Yitzhak Gadre, Gabriel Ilharco, Alex Fang, Jonathan Hayase, Georgios
  Smyrnis, Thao Nguyen, Ryan Marten, Mitchell Wortsman, Dhruba Ghosh, Jieyu
  Zhang, et~al.
\newblock Datacomp: In search of the next generation of multimodal datasets.
\newblock \emph{arXiv preprint arXiv:2304.14108}, 2023.

\bibitem[Google(2023)]{bard}
Google.
\newblock Bard.
\newblock \url{https://bard.google.com}, 2023.
\newblock Accessed: 2023-10-09.

\bibitem[Gupta \& Kembhavi(2023)Gupta and Kembhavi]{gupta2023visual}
Tanmay Gupta and Aniruddha Kembhavi.
\newblock Visual programming: Compositional visual reasoning without training.
\newblock In \emph{Proceedings of the IEEE/CVF Conference on Computer Vision
  and Pattern Recognition}, pp.\  14953--14962, 2023.

\bibitem[Kendall \& Stuart(1973)Kendall and Stuart]{kendall1973advanced}
G~Kendall and A~Stuart.
\newblock The advanced theory of statistics: inference and relationship.
  griffin, 1973.

\bibitem[Kendall(1938)]{kendall1938new}
Maurice~G Kendall.
\newblock A new measure of rank correlation.
\newblock \emph{Biometrika}, pp.\  81--93, 1938.

\bibitem[Koh et~al.(2023)Koh, Fried, and Salakhutdinov]{koh2023generating}
Jing~Yu Koh, Daniel Fried, and Ruslan Salakhutdinov.
\newblock Generating images with multimodal language models.
\newblock \emph{arXiv preprint arXiv:2305.17216}, 2023.

\bibitem[Lauren{\c{c}}on et~al.(2023)Lauren{\c{c}}on, Saulnier, Tronchon,
  Bekman, Singh, Lozhkov, Wang, Karamcheti, Rush, Kiela,
  et~al.]{laurenccon2023obelisc}
Hugo Lauren{\c{c}}on, Lucile Saulnier, L{\'e}o Tronchon, Stas Bekman, Amanpreet
  Singh, Anton Lozhkov, Thomas Wang, Siddharth Karamcheti, Alexander~M Rush,
  Douwe Kiela, et~al.
\newblock Obelisc: An open web-scale filtered dataset of interleaved image-text
  documents.
\newblock \emph{arXiv preprint arXiv:2306.16527}, 2023.

\bibitem[Li et~al.(2020)Li, Kong, and Zhou]{li2020improved}
Chunye Li, Liya Kong, and Zhiping Zhou.
\newblock Improved-storygan for sequential images visualization.
\newblock \emph{Journal of Visual Communication and Image Representation}, pp.\
   102956, 2020.

\bibitem[Li et~al.(2023)Li, Gan, Yang, Yang, Li, Wang, and
  Gao]{li2023multimodal}
Chunyuan Li, Zhe Gan, Zhengyuan Yang, Jianwei Yang, Linjie Li, Lijuan Wang, and
  Jianfeng Gao.
\newblock Multimodal foundation models: From specialists to general-purpose
  assistants.
\newblock \emph{arXiv preprint arXiv:2309.10020}, 2023.

\bibitem[Li et~al.(2019)Li, Gan, Shen, Liu, Cheng, Wu, Carin, Carlson, and
  Gao]{li2019storygan}
Yitong Li, Zhe Gan, Yelong Shen, Jingjing Liu, Yu~Cheng, Yuexin Wu, Lawrence
  Carin, David Carlson, and Jianfeng Gao.
\newblock Storygan: A sequential conditional gan for story visualization.
\newblock In \emph{CVPR}, 2019.

\bibitem[Lin et~al.(2023)Lin, Chen, Pathak, Zhang, and
  Ramanan]{lin2023visualgptscore}
Zhiqiu Lin, Xinyue Chen, Deepak Pathak, Pengchuan Zhang, and Deva Ramanan.
\newblock Visualgptscore: Visio-linguistic reasoning with multimodal generative
  pre-training scores.
\newblock \emph{arXiv preprint arXiv:2306.01879}, 2023.

\bibitem[Liu et~al.(2023{\natexlab{a}})Liu, Wu, Zhong, Zhang, and
  Xie]{liu2023intelligent}
Chang Liu, Haoning Wu, Yujie Zhong, Xiaoyun Zhang, and Weidi Xie.
\newblock Intelligent grimm--open-ended visual storytelling via latent
  diffusion models.
\newblock \emph{arXiv preprint arXiv:2306.00973}, 2023{\natexlab{a}}.

\bibitem[Liu et~al.(2023{\natexlab{b}})Liu, Li, Wu, and Lee]{liu2023visual}
Haotian Liu, Chunyuan Li, Qingyang Wu, and Yong~Jae Lee.
\newblock Visual instruction tuning.
\newblock \emph{arXiv preprint arXiv:2304.08485}, 2023{\natexlab{b}}.

\bibitem[Liu et~al.(2023{\natexlab{c}})Liu, Iter, Xu, Wang, Xu, and
  Zhu]{liu2023gpteval}
Yang Liu, Dan Iter, Yichong Xu, Shuohang Wang, Ruochen Xu, and Chenguang Zhu.
\newblock Gpteval: Nlg evaluation using gpt-4 with better human alignment.
\newblock \emph{arXiv preprint arXiv:2303.16634}, 2023{\natexlab{c}}.

\bibitem[Maharana \& Bansal(2021)Maharana and Bansal]{maharana2021integrating}
Adyasha Maharana and Mohit Bansal.
\newblock Integrating visuospatial, linguistic and commonsense structure into
  story visualization.
\newblock \emph{arXiv preprint arXiv:2110.10834}, 2021.

\bibitem[Maharana et~al.(2021)Maharana, Hannan, and
  Bansal]{maharana2021improving}
Adyasha Maharana, Darryl Hannan, and Mohit Bansal.
\newblock Improving generation and evaluation of visual stories via semantic
  consistency.
\newblock \emph{arXiv preprint arXiv:2105.10026}, 2021.

\bibitem[Maharana et~al.(2022)Maharana, Hannan, and
  Bansal]{maharana2022storydall}
Adyasha Maharana, Darryl Hannan, and Mohit Bansal.
\newblock Storydall-e: Adapting pretrained text-to-image transformers for story
  continuation.
\newblock In \emph{ECCV}, 2022.

\bibitem[Meng et~al.(2021)Meng, He, Song, Song, Wu, Zhu, and
  Ermon]{meng2021sdedit}
Chenlin Meng, Yutong He, Yang Song, Jiaming Song, Jiajun Wu, Jun-Yan Zhu, and
  Stefano Ermon.
\newblock Sdedit: Guided image synthesis and editing with stochastic
  differential equations.
\newblock \emph{arXiv preprint arXiv:2108.01073}, 2021.

\bibitem[Microsoft(2023)]{bingchat}
Microsoft.
\newblock Bingchat.
\newblock \url{https://www.microsoft.com/en-us/edge/features/bing-chat}, 2023.
\newblock Accessed: 2023-10-09.

\bibitem[OpenAI(2023{\natexlab{a}})]{chatgpt}
OpenAI.
\newblock Chatgpt, 2023{\natexlab{a}}.

\bibitem[OpenAI(2023{\natexlab{b}})]{gpt4v}
OpenAI.
\newblock Gpt-4v(ision) system card.
\newblock 2023{\natexlab{b}}.
\newblock URL \url{https://cdn.openai.com/papers/GPTV_System_Card.pdf}.

\bibitem[OpenAI(2023{\natexlab{c}})]{openai2023gpt4}
OpenAI.
\newblock Gpt-4 technical report, 2023{\natexlab{c}}.

\bibitem[Pan et~al.(2022)Pan, Qin, Li, Xue, and Chen]{pan2022synthesizing}
Xichen Pan, Pengda Qin, Yuhong Li, Hui Xue, and Wenhu Chen.
\newblock Synthesizing coherent story with auto-regressive latent diffusion
  models.
\newblock \emph{arXiv preprint arXiv:2211.10950}, 2022.

\bibitem[Podell et~al.(2023)Podell, English, Lacey, Blattmann, Dockhorn,
  M{\"u}ller, Penna, and Rombach]{podell2023sdxl}
Dustin Podell, Zion English, Kyle Lacey, Andreas Blattmann, Tim Dockhorn, Jonas
  M{\"u}ller, Joe Penna, and Robin Rombach.
\newblock Sdxl: Improving latent diffusion models for high-resolution image
  synthesis.
\newblock \emph{arXiv preprint arXiv:2307.01952}, 2023.

\bibitem[Radford et~al.(2021)Radford, Kim, Hallacy, Ramesh, Goh, Agarwal,
  Sastry, Askell, Mishkin, Clark, et~al.]{radford2021learning}
Alec Radford, Jong~Wook Kim, Chris Hallacy, Aditya Ramesh, Gabriel Goh,
  Sandhini Agarwal, Girish Sastry, Amanda Askell, Pamela Mishkin, Jack Clark,
  et~al.
\newblock Learning transferable visual models from natural language
  supervision.
\newblock In \emph{ICML}, 2021.

\bibitem[Rombach et~al.(2022)Rombach, Blattmann, Lorenz, Esser, and
  Ommer]{rombach2022high}
Robin Rombach, Andreas Blattmann, Dominik Lorenz, Patrick Esser, and Bj{\"o}rn
  Ommer.
\newblock High-resolution image synthesis with latent diffusion models.
\newblock In \emph{CVPR}, 2022.

\bibitem[Shen et~al.(2023)Shen, Song, Tan, Li, Lu, and
  Zhuang]{shen2023hugginggpt}
Yongliang Shen, Kaitao Song, Xu~Tan, Dongsheng Li, Weiming Lu, and Yueting
  Zhuang.
\newblock Hugginggpt: Solving ai tasks with chatgpt and its friends in
  huggingface.
\newblock \emph{arXiv preprint arXiv:2303.17580}, 2023.

\bibitem[Song et~al.(2020)Song, Rui~Tam, Chen, Lu, and
  Shuai]{song2020character}
Yun-Zhu Song, Zhi Rui~Tam, Hung-Jen Chen, Huiao-Han Lu, and Hong-Han Shuai.
\newblock Character-preserving coherent story visualization.
\newblock In \emph{ECCV}, 2020.

\bibitem[Sun et~al.(2023)Sun, Yu, Cui, Zhang, Zhang, Wang, Gao, Liu, Huang, and
  Wang]{sun2023generative}
Quan Sun, Qiying Yu, Yufeng Cui, Fan Zhang, Xiaosong Zhang, Yueze Wang,
  Hongcheng Gao, Jingjing Liu, Tiejun Huang, and Xinlong Wang.
\newblock Generative pretraining in multimodality.
\newblock \emph{arXiv preprint arXiv:2307.05222}, 2023.

\bibitem[Sur{\'\i}s et~al.(2023)Sur{\'\i}s, Menon, and
  Vondrick]{suris2023vipergpt}
D{\'\i}dac Sur{\'\i}s, Sachit Menon, and Carl Vondrick.
\newblock Vipergpt: Visual inference via python execution for reasoning.
\newblock \emph{arXiv preprint arXiv:2303.08128}, 2023.

\bibitem[Sz{\H{u}}cs \& Al-Shouha(2022)Sz{\H{u}}cs and
  Al-Shouha]{szHucs2022modular}
G{\'a}bor Sz{\H{u}}cs and Modafar Al-Shouha.
\newblock Modular storygan with background and theme awareness for story
  visualization.
\newblock In \emph{International Conference on Pattern Recognition and
  Artificial Intelligence}, pp.\  275--286, 2022.

\bibitem[Wu et~al.(2023)Wu, Yin, Qi, Wang, Tang, and Duan]{wu2023visual}
Chenfei Wu, Shengming Yin, Weizhen Qi, Xiaodong Wang, Zecheng Tang, and Nan
  Duan.
\newblock Visual chatgpt: Talking, drawing and editing with visual foundation
  models.
\newblock \emph{arXiv preprint arXiv:2303.04671}, 2023.

\bibitem[Yang et~al.(2023)Yang, Li, Lin, Wang, Lin, Liu, and
  Wang]{yang2023dawn}
Zhengyuan Yang, Linjie Li, Kevin Lin, Jianfeng Wang, Chung-Ching Lin, Zicheng
  Liu, and Lijuan Wang.
\newblock The dawn of lmms: Preliminary explorations with gpt-4v (ision).
\newblock \emph{arXiv preprint arXiv:2309.17421}, 2023.

\bibitem[Yang* et~al.(2023)Yang*, Li*, Wang*, Lin*, Azarnasab*, Ahmed*, Liu,
  Liu, Zeng, and Wang]{yang2023mmreact}
Zhengyuan Yang*, Linjie Li*, Jianfeng Wang*, Kevin Lin*, Ehsan Azarnasab*,
  Faisal Ahmed*, Zicheng Liu, Ce~Liu, Michael Zeng, and Lijuan Wang.
\newblock Mm-react: Prompting chatgpt for multimodal reasoning and action.
\newblock \emph{arXiv preprint arXiv:2303.11381}, 2023.

\bibitem[Zeng et~al.(2019)Zeng, Li, and Zhang]{zeng2019pororogan}
Gangyan Zeng, Zhaohui Li, and Yuan Zhang.
\newblock Pororogan: An improved story visualization model on pororo-sv
  dataset.
\newblock In \emph{Proceedings of the International Conference on Computer
  Science and Artificial Intelligence}, 2019.

\bibitem[Zhang \& Agrawala(2023)Zhang and Agrawala]{zhang2023adding}
Lvmin Zhang and Maneesh Agrawala.
\newblock Adding conditional control to text-to-image diffusion models.
\newblock \emph{arXiv preprint arXiv:2302.05543}, 2023.

\bibitem[Zhu et~al.(2023)Zhu, Hessel, Awadalla, Gadre, Dodge, Fang, Yu,
  Schmidt, Wang, and Choi]{zhu2023multimodal}
Wanrong Zhu, Jack Hessel, Anas Awadalla, Samir~Yitzhak Gadre, Jesse Dodge, Alex
  Fang, Youngjae Yu, Ludwig Schmidt, William~Yang Wang, and Yejin Choi.
\newblock Multimodal c4: An open, billion-scale corpus of images interleaved
  with text.
\newblock \emph{arXiv preprint arXiv:2304.06939}, 2023.

\end{thebibliography}
\bibliographystyle{iclr2024_conference}

\end{document}